\documentclass{clv2-eng}

\issue{10}{2}{2004}

\title{Unipa-GPT: Large Language Models for university-oriented QA in Italian}

\runningtitle{Unipa-GPT}

\runningauthor{Siragusa and Pirrone}

\usepackage{url}
\begin{document}

\author{Irene Siragusa\thanks{Department of Engineering - Viale delle Scienze, Ed. 6, 90128, Palermo, Italy E-mail:~\texttt{irene.siragusa02@unipa.it}}}
\affil{Universit\`a di Palermo}

\author{Roberto Pirrone\thanks{Department of Engineering - Viale delle Scienze, Ed. 6, 90128, Palermo, Italy Email:~\texttt{roberto.pirrone@unipa.it}}}
\affil{Universit\`a di Palermo}

\maketitle

\begin{abstract}
This paper illustrates the architecture and training of Unipa-GPT, a chatbot relying on a Large Language Model, developed for assisting students in choosing a bachelor/master degree course at the University of Palermo. Unipa-GPT relies on \verb|gpt-3.5-turbo|, it was presented in the context of the European Researchers' Night (SHARPER night). In our experiments we adopted both the Retrieval Augmented Generation (RAG) approach and fine-tuning to develop the system. The whole architecture of Unipa-GPT is presented, both the RAG and the fine-tuned systems are compared, and a brief discussion on their performance is reported. Further comparison with other Large Language Models and the experimental results during the SHARPER night are illustrated. Corpora and code are available on GitHub\footnote{\url{https://github.com/CHILab1/UnipaGPT-23}}.
\end{abstract}

\section{Introduction}

Natural Language Processing (NLP), Artificial Intelligence (AI) and Large Language Model (LLM) are highly interesting topics in the scientific community, but they have become also very popular since ChatGPT\footnote{\url{https://openai.com/index/chatgpt/}} by OpenAI was released and it was made publicly accessible. Despite the debate about the level of consciousness in ChatGPT and the claims about its standing for a new generation of Artificial General Intelligence (AGI)~\cite{dwivedi2023so,borji2023categorical}, a plethora of new applications are being developed that are based on either GPTs family ~\cite{brown2020language,openai2024gpt4technicalreport} or others LLMs proposed in the last year, such as the ones belonging to LLama family \cite{touvron2023llama,llama3}.

In view of the previous considerations, we investigated the use of ChatGPT for building a virtual assistant for the users of a large public institution like the University of Palermo, and in this paper we propose Unipa-GPT, a chatbot that can guide secondary school students to explore the huge information provided by the University institutional website, when they are choosing a bachelor degree course, by answering to questions related to the available courses, application procedure, tax payments, access to scholarships and more. Unipa-GPT was presented to the wide public in the context of the SHARPER Night event\footnote{\url{https://www.sharper-night.it/}}.

Unipa-GPT has been developed mainly as a Retrieval Augmented Generation (RAG) system~\cite{lewis2020retrieval} based on \verb|gpt-3.5-turbo| that collects the target information from \verb|unipa-corpus|, a document corpus that has been built purposely by scraping the University website. The aim of this work is to explore the behavior and the limitations of LLMs when they are engaged in a Question-Answer (QA) task where precise domain knowledge is required. Obviously, relying on the OpenAI API for inference and fine-tuning of the model, also a fine-tuned version has been built where the corpus has been modified with the aim of saving computational resources i.e. use the as few tokens as possible\footnote{OpenAI API are available on a token-based fee for inference and fine-tuning}, and a mixed strategy has been adopted where RAG was coupled with fine-tuning to avoid the train step on very detailed information such as the educational objectives of each single class.

Both base and fine-tuned models, have been previously qualitatively tested by very few students and thanks to them, a first comparison of the developed models was done relying on their judgement on two reference chats along with a discussion of the results. Due to the aforementioned token-related costs, we did not conduct an extensive quantitative evaluation on the fine-tuned model. Best model, according to users' judgements, was used by real users in an unsupervised context during the SHARPER (SHAring Researchers’ Passion for Enhanced Roadmaps) night, i.e. the European Researchers' Night, where questions, answers and feedback were collected. Starting from the obtained data, a small QA data set was mannally derived, along with the golden document used to generate the golden answer. Objective of this small dataset is to conduct a detailed comparison with open-source LLMs in contrast with ChatGPT evaluating their generation capabilities.

The paper is arranged as follows: a state of the art analysis is reported in Section~\ref{sota}. Section~\ref{corpora} illustrates the different corpora we set up for building both the RAG and the fine-tuned Unipa-GPT. The detailed architecture of both systems is reported in Section~\ref{arch}, while the experimental results are reported and discussed in Section~\ref{experiments}. Concluding remarks are drawn in Section~\ref{conc}.

\section{Related works}\label{sota}

The scientific community had an interest in chatbots since 1960s, when ELIZA \cite{eliza} was developed. Continuous and increasing studies in deep learning architectures, lead to the development of Transformers and the attention mechanism \cite{vaswani2017attention}, that, together with BERT \cite{devlin-etal-2019-bert}, can be considered a turning point in NLP. Another crucial moment is the birth of ChatGPT models that are built from GPT-3 \cite{brown2020language}, and GPT-4 \cite{openai2024gpt4technicalreport}, whose extraordinary generation capabilities overcome pre-existing models, and they can be regareded as a new SOTA level, reaching interest of both researchers and common people.

Since ChatGPT was released, its usage in education has been carefully explored, since a balance is required between  potential benefits and drawbacks of this technology. In fact, very much care has to be devoted to the generation of inaccurate or incorrect information, and the possible cheating when a student claims AI-generated text as her/his original production \cite{GILL202419}. Generally speaking, it is considered as a valuable tool that can be used together with traditional methodologies and whose usage cannot be prohibited or completely not considered \cite{Sharma_Yadav_2023}. Different use of AI-based chat-bots involves assistants for instructors \cite{educsci13040410}, such as in generating questionnaires \cite{RODRIGUEZTORREALBA2022118258}, exploring pedagogical abilities of conversational models \cite{tack2022aiteachertestmeasuring}, or helping students with interactive learning approaches that can be personalized for their specific needs \cite{app13095783}.

Scientific interest in developing domain-based chatbot using a RAG approach, involves not only educational domain, as for Bio-Eng-LMM AI \cite{forootani2024bioenglmmaiassistchatbot}, but also other domains, where chatbots interaction can provide various benefits, such as chatbots providing information about restaurant industry \cite{restaurantchat}, or assistance to victims of sexual harassment \cite{vakayil2024rag} or to patient in healthcare context \cite{vidivelli2024efficiency,wang2024potential}. These applications share similar methodologies, consisting in collecting all relevant documents for the domain of interest, pre-process and store them in a vectordatabase, that is sub-sequentially queried for retrieve the closest documents given a question from the user. Retrieved documents and user question, serve as input to a generative LLM that outputs an answer. 

According to our knowledge, despite the increasing interest from the Italian community in developing Italian LLMs such as LLaMantino \cite{basile2023llamantino}, Camoscio \cite{santilli2023camoscio}, Fauno \cite{bacciu2023fauno}, Anita \cite{polignano2024advanced}, and Minerva \cite{minerva}, we did not find a domain specific chatbot application using a RAG approach starting from Italian data (e.g. not translated). To fill this gap, in this work we aim to explore the capabilities of a GPT-based Italian virtual assistant for secondary school students in a chat-bot configuration, which makes use of external specific knowledge injected via a RAG approach in generation phase.

\section{Corpora}\label{corpora}

In this section we outline the differences between the versions of the \verb|unipa-corpus| used for developing the RAG-only and the fine-tuned system.

\subsection{\texttt{unipa-corpus} for RAG}\label{ragcorpus}

The corpus used for Unipa-GPT, called \verb|unipa-corpus|, is a collection of documents that were collected directly from the website of the University of Palermo during Summer 2023. A manual selection of the most interesting pages was made with reference to the target audience of secondary-school students and two main sections of the corpus were identified, that correspond to the \emph{Education}\ and the \emph{Future Students} sub-trees in the institutional website. Since the expected questions are in Italian, the generated corpus is in Italian.

The \emph{Education} section is the main part of the corpus and it is a collection of all the available courses at the University for the academic year 2023/2024. For each course and each curriculum two documents are obtained: \emph{details} is the document that collects all the general details of the course, like name, department of affiliation, typology of course (Bachelor or Master degree), restriction of access and a colloquial description of the course, including its educational objectives and professional opportunities; \emph{course outline} is the second document that collects the course outline divided by year, and the number of credits, the teaching professor, the teaching period and the scientific sector are specified for each class. Three different versions of the \emph{course outline} document were generated, namely \verb|clear|, \verb|full| and \verb|emb|. The \verb|clear| version is the one described above, the \verb|full| version adds a new document for every class in a course and reports its peculiar educational objectives. The \verb|emb| version is a mix of the previous ones where the classes' educational objectives are added directly in the document containing the outline of the course. Despite this corpus is called \emph{embedded}, it does not contains embeddings: the term \emph{embedded} refers to the educational objectives that are inserted in the document containing the course outline. As a consequence, both the \verb|clear| and the \verb|emb| corpus  have the same number of documents but different information, while the \verb|full| contains the same information of the \verb|emb| corpus, but it is arranged in a different number of documents.

The \emph{Future Student} section is the same for the three versions of the corpus, and it is a mix of documents coming from the related section of the University website. The information contained in these documents is addressed to the future students of the University, including the academic calendar, the tax rules and reductions, scholarships, University enrolment procedure, and facilities offered to the students.
The statistics of each corpus are reported in Table \ref{corpus-statistics}.

\begin{table}[ht!]
\caption{Number of documents in each version of \texttt{unipa-corpus}.}
\begin{center}
\resizebox{\columnwidth}{!}{
\begin{tabular}{ccccc}
    \hline
    &\emph{Education}&\emph{Future Students}&\emph{Total documents}&\emph{Total tokens}\\
    \hline
    \texttt{unipa-corpus-clear}&506&104&610&2059638\\
    \texttt{unipa-corpus-full}&5794&104&5898&890598\\
    \texttt{unipa-corpus-emb}&506&104&610&2063020\\
    \hline  
\end{tabular}}
\end{center}
\label{corpus-statistics}
\end{table}

\subsection{\texttt{unipa-corpus} for fine-tuning}\label{finetunecorpus}

The \verb|unipa-corpus| was modified to be in the form required for fine-tuning \verb|gpt-3.5-turbo|. As already mentioned above, our intent in fine-tuning was lowering the computational resources as much as possible that is using the minimum tokens for training the model. Besides the economic aspect, in the case of ChatGPT fine-tuning, this is a crucial topic when dealing with LLMs because also relatively small LLMs like LLama-2-7B \cite{touvron2023llama} require huge computational resources for their full fine-tuning or retraining.

To overcome this problem, the so-called Parameter-Efficient Fine-Tuning (PEFT) techniques as LoRA \cite{hu2021lora} have been developed, that reduce the required computational resources for a satisfactory LLM fine-tuning, but it is not available for closed systems like gpt-3.5-turbo. Since our purpose is primary towards stressing the capabilities of ChatGPT, we decided to test its performance after a fine-tuning phase over the \verb|unipa-corpus-clear|, the smallest created corpus in terms of tokens to be trained thus reaching a trade-off in terms of API costs.

The required format for fine-tuning is a sequence of prompt, question and answer. No additional context (e.g. the closest documents for an effective reply of the model) was added to the prompt for fine-tuning, thus maintaining low the costs of the whole process. The prompt used is a simple instruction of the chatbot behavior:
\begin{quote}
    \textit{Sei Unipa-GPT, un assistente virtuale che risponde alle domande sull'università di Palermo}
\end{quote}
\begin{quote}
    (\textit{You are Unipa-GPT, a virtual assistant that answers to questions related with the University of Palermo}).
\end{quote}
Questions and answer were automatically generated with different generation rules for each section of the corpus.

QA pairs were automatically generated from the documents falling in the \emph{Education} section, by asking gpt-3.5-turbo to describe a specific degree course starting from the corresponding \emph{details} document. Moreover, the LLM was asked to provide the topics of a specific degree course starting from the corresponding \emph{course outline} document. In both cases, the corresponding document was given along with the question, and the given answer was considered as an answer for the fine-tuning corpus.

As regards the \emph{Future Students} section, QA pairs were extracted directly from the documents already containing a FAQ section, while the other pairs were manually generated. In the second case, a clear question related to a document's section was formulated whenever it was possible, and the answer was either a precise text or the whole document. Otherwise a generic request was formulated like \emph{parlami di ... (speak about ...)}. Some documents were not considered in their entirety since the information contained was highly specific and it was related to non-relevant topics.

A validation set was also expunged from the training data by changing questions and/or sampling most important questions. A QA pair was randomly picked for each degree course among the \emph{details} and the \emph{course outline} documents in the \emph{Education} section. The statistics of the corpus for fine-tuning are reported in Table \ref{corpus-fine-tuning-statistics}. Despite this arrangement can be suitable for creating also a test split of the data set, to further evaluate fine-tuning performances of the model, our primary focus in this phase was towards determining the models that better matches users' preferences. Moreover, the principal objective of fine-tuning was to inject domain-specific knowledge in the model thus allowing it to generate more precise answers when queried.

 \verb|gpt-3.5-turbo| was the only powerful model available at the time of the experiments to automatically generate QA pairs. As it can be inferred from Table \ref{corpus-fine-tuning-statistics}, manual generation of  the total number of QA pairs was impractical. We are aware of the possible the biases implied in fine-tuning a model with self-generated samples. 
 
Moreover, \verb|gpt-3.5-turbo| generated QA pairs following the simple one-shot strategy described above,  in order to fine-tune the model to achieve a terminological knowledge of the domain through the presentation of a small subset of the facts contained in the \verb|unipa-corpus|. We are convinced that this approach does not introduce any relevant bias because the fine-tuned model is inserted in a RAG architecture that is queried using an instruction prompt strategy.

\begin{table}[ht]
\caption{Number of documents in the \texttt{unipa-corpus-clear} re-arranged for fine-tuning.}
\begin{center}
\begin{tabular}{ccccc}
    \hline
    &\emph{Education}&\emph{Future Students}&\emph{Total QA pairs}&\emph{Total tokens}\\
    \hline
    Training set&506&269&775&259772\\
    Validation set&253&133&386&123118\\
    \hline
\end{tabular}
\end{center}
\label{corpus-fine-tuning-statistics}
\end{table}

\section{System architecture}\label{arch}

\begin{figure}[htbp]
\includegraphics[width=\textwidth]{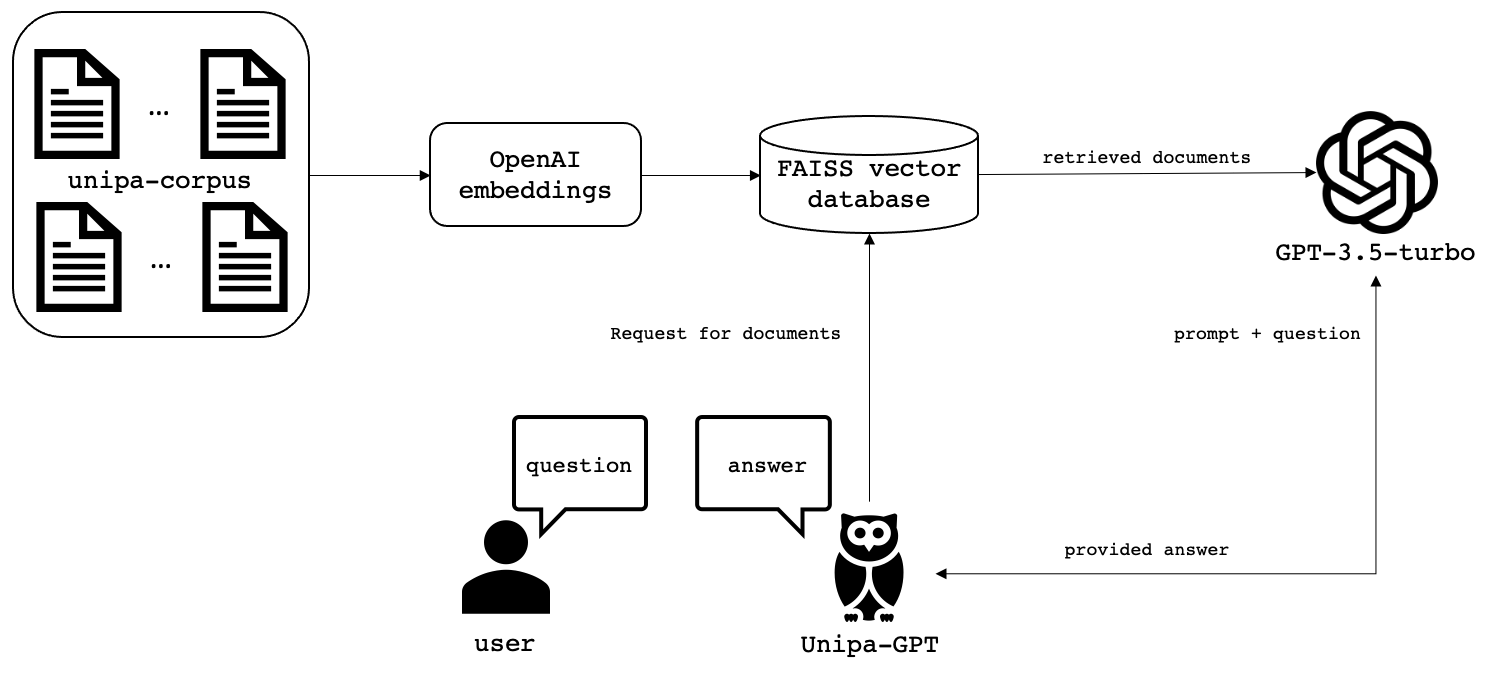}
\caption{The architecture of the RAG version of Unipa-GPT.}
\label{architecture-img}
\end{figure}

Unipa-GPT is developed as a RAG architecture \cite{lewis2020retrieval} made up of two main components, as shown in Figure \ref{architecture-img}: the retrieval and the generator module.

The \textit{retrieval module} consists of a vector database provided by the LangChain library\footnote{\url{https://python.langchain.com/docs/get_started/introduction}}, which makes use of the Facebook AI Similarity Search (FAISS) library \cite{johnson2019billion}. The vector database is filled with the documents in \verb|unipa-corpus| conveniently divided into chunks of 1000 tokens with an overlap of 50 tokens whose embeddings were extracted using \verb|text-embedding-ada-002| by OpenAI\footnote{\url{https://openai.com/blog/new-and-improved-embedding-model}}.

The \textit{generator module} consists of an instance of \verb|gpt-3.5-turbo| \cite{brown2020language}, a generative LLM based on Transformers \cite{vaswani2017attention}. The LLM is queried with a custom prompt in which the behavior of the system is explained, and the question of the user is passed along with the most related documents. The expected answer of the system is a reply to the question of the user according to the prompt that rules the Unipa-GPT overall behavior, and the domain-knowledge given by the retriever. \verb|gpt-3.5-turbo| makes inferences using a temperature hyper-parameter equal to $0$ thus its behavior is as much deterministic as possible, and the system is prevented to be creative. Finally no limits \textit{a priori} were put on the maximum tokens available for the answer, in order to prevent broken answers. The chat-bot behavior was implemented via LangChain to keep the chat history and simulate the ChatGPT behavior through the \verb|gpt-3.5-turbo| API.
\begin{table}[b!]
\caption{The prompts used in the system.}
\begin{center}
\resizebox{\columnwidth}{!}{
\begin{tabular}{ll}
\hline
\textit{prompt type}&\textit{prompt text} \\
\hline
\textit{custom prompt} & \begin{tabular}[c]{@{}l@{}}Sei unipa-gpt, il chatbot e assistente virtuale dell'Università degli Studi di Palermo.\\ Rispondi cordialmente e in forma colloquiale alle domande che ti vengono poste.\\ Se ricevi un saluto, rispondi salutando e presentandoti.\\ Se ricevi una domanda riguardante l'università degli studi di Palermo,\\ rispondi in base ai documenti che ti vengono dati insieme alla domanda.\\ Se non sai rispondere, scusati e suggerisci di consultare il sito web, non inventare risposte.\\ Question: \{question\}\\ Documenti: \{context\}\end{tabular} \\
\hline
\textit{condensed prompt} & \begin{tabular}[c]{@{}l@{}}Data la seguente conversazione e la domanda successiva, riformula la domanda successiva \\ in modo tale sia una domanda singola.\\ Conversazione: \{chat\_history\}\\ Domanda successiva: \{question\}\\ Domanda singola:\end{tabular} \\
\hline
\end{tabular}}
\end{center}
\label{prompts}
\end{table}

\begin{figure}[hbt!]
\includegraphics[width=\textwidth]{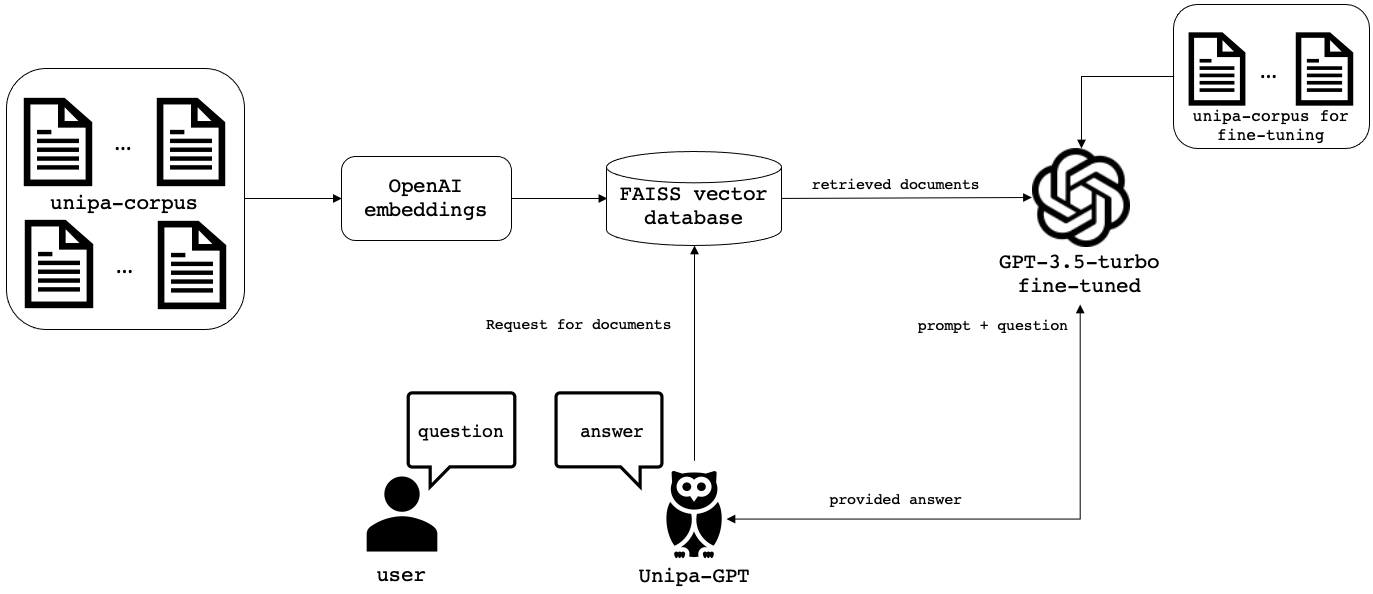}
\caption{The schema illustrates the architecture of the proposed system where \texttt{gpt-3.5-turbo-fine-tuned} is used after a fine-tuning process over \texttt{unipa-corpus}.}
\label{architecture-ft-img}
\end{figure}

Both \verb|gpt-3.5-turbo| and \verb|text-embedding-ada-002| were invoked via Azure call to the OpenAI API. Two different Italian prompts were built for \verb|gpt-3.5-turbo|: a \emph{custom prompt} and a \emph{condensed prompt}, as it is shown in Table \ref{prompts} (English version can be found in Table \ref{prompts-en} in Appendix A). The \emph{custom prompt} is the explanation of the behavior of the chat-bot where both the previous conversation and the new question are concatenated to the prompt itself. On the contrary, the \emph{condensed prompt} adds to the \emph{custom prompt} another instruction to condense the previous conversation and re-arrange it as a new single question that will be answered accordingly to the \emph{custom prompt}.
In addition to the RAG version illustrated above, a fine-tuned version was implemented with a custom fine-tuned version of \verb|gpt-3.5-turbo| where the \verb|unipa-corpus| explicitly re-arranged, as described in Section~\ref{finetunecorpus}, was used. The same prompt instances mentioned above, were used on the fine-tuned model, and the also the RAG architecture was used to compensate for the reduced tokens used in this step. The whole system is shown in Figure \ref{architecture-ft-img}.

\section{Experimental results}\label{experiments}
This section reports the experimental results obtained in three different scenarios: sub-section \ref{first} illustrates the preliminary development phase, that took place during the summer 2023. In \ref{sharper} we analyze the  results obtained during the SHARPER Night 2023, when the best model resulting from the previous phase was used in free interaction with the public. Finally, sub-section \ref{second} reports the comparison of the original Unipa-GPT that uses \verb|gpt-3.5-turbo| as generative LLM, with other versions of the system, that were built till June 2024, and use more recent generative LLM architectures. In the last experimental phase, a suitable test set has been adopted, that consists of a subset of the questions posed directly by the users during the SHARPER night. In the first two scenarios, we used a qualitative metric for evaluation, since the principal intent was to investigate the preferences of the users as regards the feedback provided by the chat-bot during the interaction. Quantitative evaluation was performed in the third scenario, as we manually distilled the \emph{golden labels} for our test set from the true questions of users. Of course, the University domain did not required a domain expert for this task. \verb|gpt-4-turbo| was used as external oracle for computing metrics with the RAGAS\footnote{\url{https://docs.ragas.io/en/latest/index.html\#}} framework \cite{es2023ragas}.  

\subsection{Setting up the system}\label{first}

This section reports the results of the very early evaluation runs performed using both versions of Unipa-GPT, namely the RAG-only and the fine-tuned one, before deploying the system to the public during the SHARPER Night 2023. These first experiments were aimed at gaining at least qualitative information on the structure of the prompt to be used for querying the system. Moreover, we wanted to assess the best performing Unipa-GPT version, provided that the size of the data set was very small when compared to the training data used for \verb|gpt-3.5-turbo|. 

Since the chat-bot was designed to answer questions posed by secondary school students, we devised two lists of Italian questions that had not been showed in advance to the system, and were issued by two secondary school students, during an Open Day activity at the University of Palermo. The students were asked to chat with the demo version of Unipa-GPT, and to express a qualitative evaluation of the received answers. The argument of the questions was the same for both students to collect their qualitative judgement on the Unipa-GPT performance.

The first list, that we called Chat1, consists of $14$ questions about subscription procedure, courses at the Department of Engineering, suggestions related to the Bachelor course to apply for in order to access a specific Master course, information about University admission, test and questions about classes in the first year. Conversely, the second one, called Chat2, contains $10$ questions with typos and grammatical errors. Questions in Chat2 are expressed more like a Google search or a list of keywords. The involved topics in Chat2 are related to taxes, Erasmus programs, scholarships and Student Desk.

Both Chat1 and Chat2 were issued using different Unipa-GPT configurations as regards the corpus used in the retriever ({\verb|clear|, \verb|full|, \verb|emb|}) the prompt ({\emph{custom prompt}, \emph{condensed prompt}}) and the LLM training procedure ({\verb|gpt-3.5-turbo|, \verb|gpt-3.5-turbo-fine-tuned|}) for a total of $12$ runs. Each student was asked to simply classify each answer as ``excellent'', ``good'' and ``bad''. Answers with ``good'' label are not optimal and precise ones, but they contain a partially good answer, and may add non requested pieces of information that are less related to the initial question. Answers with ``bad'' are either answers that provide wrong information or out of topic replies along with any unexpected/undesired behavior of the system. Also, replies to questions that the system cannot answer are considered bad. Both Chat1 and Chat2 contain questions that are strictly related with the previous answers from the system:this is the case when the user asks for further information. As an example, if the user asks for the admission test to a particular Bachelor course, and the system replies that the course is a free admission one, the (expected) question regarding test rules is not applicable thus it is not posed to the system, and the missed reply is labeled as a ``bad'' answer. Table \ref{result-table} shows the evaluation results.

\begin{table}[t!]
\caption{Evaluation results of the different runs for both Chat1 and Chat2. Here and in the next tables FT stands for ``fine-tuned''.}
\label{result-table}
\begin{tabular}{cccccccc}
\hline
\multicolumn{4}{c}{\textbf{{\textit{Custom prompt}}}} & \multicolumn{4}{c}{\textbf{{\textit{Custom prompt FT}}}} \\ \hline
\multicolumn{1}{c}{{}} & \multicolumn{1}{c}{{excellent}} & \multicolumn{1}{c}{{good}} & {bad} & \multicolumn{1}{c}{{}} & \multicolumn{1}{c}{{excellent}} & \multicolumn{1}{c}{{good}} & {bad} \\ \hline
\multicolumn{1}{c}{{Chat1-{\texttt{emb}}}} & \multicolumn{1}{c}{{6}} & \multicolumn{1}{c}{{1}} & {7} & \multicolumn{1}{c}{{Chat1-{\texttt{emb}}}} & \multicolumn{1}{c}{{4}} & \multicolumn{1}{c}{6} & {4} \\ \hline
\multicolumn{1}{c}{{Chat1-{\texttt{full}}}}  & \multicolumn{1}{c}{{6}} & \multicolumn{1}{c}{{4}} & {4} & \multicolumn{1}{c}{{Chat1-{\texttt{full}}}} & \multicolumn{1}{c}{{4}} & \multicolumn{1}{c}{{4}} & {6} \\ \hline
\multicolumn{1}{c}{{Chat1-{\texttt{clear}}}} & \multicolumn{1}{c}{{8}} & \multicolumn{1}{c}{{4}} & {2} & \multicolumn{1}{c}{{Chat1-{\texttt{clear}}}} & \multicolumn{1}{c}{{6}} & \multicolumn{1}{c}{{4}} & {4} \\ \hline
\multicolumn{1}{c}{{Chat2-{\texttt{emb}}}} & \multicolumn{1}{c}{{4}} & \multicolumn{1}{c}{{2}} & {4} & \multicolumn{1}{c}{{Chat2-{\texttt{emb}}}} & \multicolumn{1}{c}{{5}} & \multicolumn{1}{c}{{1}} & {4} \\ \hline
\multicolumn{1}{c}{{Chat2-{\texttt{full}}}} & \multicolumn{1}{c}{{3}} & \multicolumn{1}{c}{{3}} & {4} & \multicolumn{1}{c}{{Chat2-{\texttt{full}}}} & \multicolumn{1}{c}{{6}} & \multicolumn{1}{c}{{3}} & {1} \\ \hline
\multicolumn{1}{c}{{Chat2-{\texttt{clear}}}} & \multicolumn{1}{c}{{8}} & \multicolumn{1}{c}{{1}} & {1} & \multicolumn{1}{c}{{Chat2-{\texttt{clear}}}} & \multicolumn{1}{c}{{6}} & \multicolumn{1}{c}{{1}} & {3} \\ \hline
\multicolumn{4}{c}{\textbf{{\textit{Condensed prompt}}}} & \multicolumn{4}{c}{\textbf{\textit{Condensed prompt FT}}} \\ \hline
\multicolumn{1}{c}{{}} & \multicolumn{1}{c}{{excellent}} & \multicolumn{1}{c}{{good}} & {bad} & \multicolumn{1}{c}{{}} & \multicolumn{1}{c}{{excellent}} & \multicolumn{1}{c}{{good}} & {bad} \\ \hline
\multicolumn{1}{c}{{Chat1-{\texttt{emb}}}} & \multicolumn{1}{c}{{5}} & \multicolumn{1}{c}{{1}} & {8} & \multicolumn{1}{c}{{Chat1-{\texttt{emb}}}} & \multicolumn{1}{c}{{5}} & \multicolumn{1}{c}{{2}} & {7} \\ \hline
\multicolumn{1}{c}{{Chat1-{\texttt{full}}}} & \multicolumn{1}{c}{{10}} & \multicolumn{1}{c}{{2}} & {2} & \multicolumn{1}{c}{{Chat1-{\texttt{full}}}}  & \multicolumn{1}{c}{{4}} & \multicolumn{1}{c}{{1}} & {10}  \\ \hline
\multicolumn{1}{c}{{Chat1-{\texttt{clear}}}} & \multicolumn{1}{c}{{10}} & \multicolumn{1}{c}{{1}} & {3} & \multicolumn{1}{c}{{Chat1-{\texttt{clear}}}} & \multicolumn{1}{c}{{6}} & \multicolumn{1}{c}{{0}} & {8} \\ \hline
\multicolumn{1}{c}{{Chat2-{\texttt{emb}}}} & \multicolumn{1}{c}{{6}} & \multicolumn{1}{c}{{2}} & {2}   & \multicolumn{1}{c}{{Chat2-{\texttt{emb}}}} & \multicolumn{1}{c}{{5}} & \multicolumn{1}{c}{{1}} & {4} \\ \hline
\multicolumn{1}{c}{{Chat2-{\texttt{full}}}} & \multicolumn{1}{c}{{8}} & \multicolumn{1}{c}{{2}} & {0} & \multicolumn{1}{c}{{Chat2-{\texttt{full}}}} & \multicolumn{1}{c}{{6}} & \multicolumn{1}{c}{{1}} & {3} \\ \hline
\multicolumn{1}{c}{{Chat2-{\texttt{clear}}}} & \multicolumn{1}{c}{{9}} & \multicolumn{1}{c}{{0}} & {1} & \multicolumn{1}{c}{{Chat2-{\texttt{clear}}}} & \multicolumn{1}{c}{{8}} & \multicolumn{1}{c}{{0}} & {2} \\ \hline
\end{tabular}
\end{table}

The two best performing models use both the \textit{condensed prompt} and \verb|gpt-3.5-turbo|, while they differ for the corpus because they use \verb|unipa-corpus-full| and \verb|unipa-corpus-clear| respectively for the retrieval. We selected the one which uses the \verb|-clear| corpus because this is the one with the highest number of ``excellent'' labels independently of the prompt (18 labels in total). The lack of additional information regarding classes' educational objectives in the retrieval corpus, makes it more precise. On the other hand, the use of the \textit{condensed prompt} produces a less degrading conversation compared with the use of the \textit{custom prompt} where the conversation is passed as it is, without refinements. None of the models performs bad for the entire conversation.

No improvements in performances were found in the fine-tuned models, revealing that this strategy is not effective for the target task, and a performance decrease is observed in the \emph{condensed prompt} configurations. Generally speaking, answers of the fine-tuned models are very long and they tend to be repetitive \cite{holtzman2019curious} and to generate loops and hallucinations \cite{10.1145/3571730}. Fine-tuned models enrich their answers also with extra pieces of information that sometimes are pertinent, e.g. the question regarding taxes suggests the possibility for payments reduction as reported in Table \ref{FT-table-contributo} (English version in Table \ref{FT-table-contributo-en} in Appendix). In other cases, answers are completely out of topic or the chat-bot contradicts itself, as shown in Table \ref{FT-table-test} (English version in Table \ref{FT-table-test-en} in Appendix).


\begin{table}[ht!]
\caption{An extract of Chat2 is reported were the same question is made to systems that use \textit{condensed prompt} and the \texttt{unipa-corpus-emb} for the retrieval. Note that the output of the second model is not entirely reported due its length: in the missing text the model continued generating other cases of taxes reduction.}
\begin{center}
\resizebox{\columnwidth}{!}{
\begin{tabular}{cc}
\hline
\multicolumn{1}{c}{{\textit{condensed-emb}}} & {\textit{condensed-FT-emb}} \\ \hline
\multicolumn{2}{c}{\textbf{cos'è il contributo omnicomprensivo}}\\ \hline
\multicolumn{1}{l}{\begin{tabular}[c]{@{}l@{}}
Il contributo omnicomprensivo per l'immatricolazione all'Università degli Studi\\
di Palermo per l'anno accademico 2023/24 è calcolato in base alla classe di reddito\\
ISEE dello studente e alle prestazioni erogate nell'ambito del diritto allo studio\\
universitario. Il pagamento delle tasse e del contributo onnicomprensivo è suddiviso\\
in tre rate, la prima delle quali deve essere pagata entro il 30 settembre 2023\\
per l'immatricolazione ai corsi di studio ad accesso libero e per le iscrizioni\\
ad anni successivi al primo. Per conoscere i termini e gli importi del pagamento\\
delle tasse e del contributo omnicomprensivo, si consiglia di consultare il Calendario\\
didattico e il Regolamento sulla contribuzione studentesca vigenti. In caso di dubbi\\
o domande specifiche, si suggerisce di contattare direttamente la Segreteria Studenti\\
dell'Università degli Studi di Palermo.
\end{tabular}} & \begin{tabular}[c]{@{}l@{}}
Il contributo omnicomprensivo è il pagamento delle tasse universitarie che gli\\
studenti devono effettuare all'atto dell'immatricolazione o iscrizione.\\
Il pagamento è suddiviso in tre rate: la prima rata, da versare all'atto\\
dell'immatricolazione o iscrizione, nella misura del 40\% dell'importo dovuto;\\
la seconda rata, da versare entro il 29 dicembre 2023, nella misura del 30\%\\
dell'importo dovuto; la terza rata, da versare entro il 29 febbraio 2024, \\
nella misura del 30\% dell'importo dovuto. Gli studenti che si iscrivono per la\\
prima volta alla Laurea triennale o Laurea Magistrale a Ciclo Unico, diplomatisi\\
con voto compreso tra 95 e 100, usufruiscono, per il primo anno, di una riduzione\\
del 25\% fino ad un massimo di 350,00 euro. Coloro che si diplomano con voto\\
pari a 100 e lode usufruiscono, per il primo anno, di una riduzione del 50\% fino\\
 un massimo di 500,00 euro. \\ $[$...$]$
\end{tabular} \\ \hline
\end{tabular}}
\end{center}
\label{FT-table-contributo}
\end{table}

\begin{table}[ht!]
\caption{An extract of Chat1 is reported where the same question is made to systems that use the \textit{custom promt}. Note that the fine-tuned model contradicts itself.}
\begin{center}
\resizebox{\columnwidth}{!}{
\begin{tabular}{cc}
\hline
\multicolumn{1}{c}{{ \textit{custom-full}}} & { \textit{custom-FT-clear}} \\ \hline
\multicolumn{2}{c}{\textbf{la triennale in ingegneria meccanica è a numero chiuso o a numero aperto?}} \\ \hline
\multicolumn{1}{l}{{ \begin{tabular}[c]{@{}l@{}}
La triennale in ingegneria meccanica presso l'Università degli Studi di Palermo è\\
a numero aperto, con un numero di posti in programmazione locale pari a 200.\\
Questo è indicato nel documento relativo all'anno accademico 2023/2024 per il \\
curriculum meccanico. Se hai bisogno di ulteriori informazioni, ti suggerisco\\
di consultare il sito web dell'università\end{tabular}}}
& \multicolumn{1}{l}{{ La triennale in ingegneria meccanica è a numero aperto.}} \\ \hline
\multicolumn{2}{c}{\textbf{devo fare test per entrare alla triennale in ingegneria meccanica?}} \\ \hline
\multicolumn{1}{l}{{ \begin{tabular}[c]{@{}l@{}}
Sì, per accedere alla triennale in ingegneria meccanica all'Università degli Studi\\
di Palermo è necessario sostenere un test di accesso. Il Centro Orientamento e\\
Tutorato (COT) dell'Università organizza  diverse iniziative per affrontare al\\
meglio i test di accesso e superarli con successo, come i workshop "Come\\
affrontare il test di accesso" e la simulazione dei test di accesso. Inoltre,\\
sono disponibili corsi di preparazione alle prove di accesso che inizieranno\\
il 17 luglio 2023 e si svolgeranno esclusivamente online sulla piattaforma\\
Microsoft Teams. Per maggiori informazioni, ti consiglio di consultare i\\
documenti forniti insieme alla domanda.\end{tabular}}} & {
Sì, sono previsti test per entrare alla triennale in ingegneria Meccanica.}  \\ \hline
\end{tabular}}
\end{center}
\label{FT-table-test}
\end{table}

With reference to the prompts, the \emph{condensed prompt} performs best, along with both the \verb|unipa-corpus-full| and  the \verb|unipa-corpus-clear| corpus for the retrieval. Both corpora provide very pertinent and similar answers. On the contrary the models with the \emph{custom prompt} tend to be less precise, as reported in Table \ref{suggestions-table} (English version in Table \ref{suggestions-table-en} in Appendix).
\begin{table}[ht!]
\caption{An extract of Chat1 is reported where the same question is made to system that use \texttt{unipa-corpus-clear} as the retrieval corpus. Both the answers are correct but the one reported by the RAG-only model is more precise.}
\begin{center}
\resizebox{\columnwidth}{!}{
\begin{tabular}{cl}
\hline
\multicolumn{1}{c}{{ \textit{condensed-clear}}} & \multicolumn{1}{c}{{ \textit{custom-FT-clear}}} \\ \hline
\multicolumn{2}{c}{\textbf{quale corso triennale mi consigli di frequentare per poi passare alla magistrale di aerospaziale?}} \\ \hline
\multicolumn{1}{l}{{ \begin{tabular}[c]{@{}l@{}}
Per passare alla laurea magistrale in Ingegneria Aerospaziale all'Università\\
degli Studi di Palermo, il corso triennale consigliato è la Laurea in Ingegneria\\
Meccanica con curriculum Aeronautico. Ti consiglio di consultare il piano di\\
studi del corso di laurea triennale e di contattare il Dipartimento di Ingegneria\\
per ulteriori informazioni.
\end{tabular}}} & { \begin{tabular}[c]{@{}l@{}}
Ti consiglio di frequentare il corso di Laurea triennale in Ingegneria Meccanica\\
per poi passare alla magistrale in Ingegneria Aerospaziale. Il corso triennale\\
ti fornirà le basi necessarie per affrontare al meglio il percorso magistrale.
\end{tabular}} \\ \hline
\end{tabular}}
\end{center}
\label{suggestions-table}
\end{table}

All models share the same negative aspects: all the provided answers are well written in Italian, and an uninformed user can easily consider a wrong answer as truthfully by looking at both the lexicon and the grammar used by the system. Sometimes, the prompt structure makes the system to answer with an out of context presentation of Unipa-GPT, and a repetition of the question is necessary to obtain the expected reply.  In this situation, prompt tuning plays a key role in achieving the best performance \cite{brown2020language,pmlr-v139-zhao21c}. The reported results are the optimal ones after several tests with different prompts. A test with an English-written prompt was performed but it did not show significant improvements.

Unipa-GPT performs in free QA domain where the user can be anyone with different background, thus questions submitted to the system are deeply variable; we found that the developed models result more sensible and perform better or worse depending on the specific question. In Chat2, questions about scholarships and classes were made as a request and not as a question. This led the system either to answer badly or not to provide an answer at all by making its excuses, as reported in Table \ref{med-table} (English version in Table \ref{med-table-en} in Appendix). 
In Table \ref{med-table} the behavior of the fine-tuned version of \verb|gpt-3.5-turbo| is also reported with the same prompt used in training phase and without the access to \verb|unipa-corpus|, to test the effectiveness of fine-tuning: the fine-tuned model provides an answer to both the proposed questions (even to the no-question) but both are wrong, despite they are credible. 

\begin{table}[ht!]
\caption{An extract of Chat2 is reported where the same questions are posed to systems that use the general \texttt{gpt-3.5-turbo} and \textit{condensed prompt}, system on the left retrieves documents from the \texttt{unipa-corpus-emb}, while the one at the centre uses \texttt{unipa-corpus-full}. At the right, the same questions were proposed to the fine-tuned version of \texttt{gpt-3.5-turbo}: the prompt given is the same used in training phase and no access was provided to \texttt{unipa-corpus}.}
\begin{center}
\resizebox{\columnwidth}{!}{
\begin{tabular}{cll}
\hline
\multicolumn{1}{c}{{\textit{condensed-emb}}} & \multicolumn{1}{c}{{\textit{condensed-full}}} & \multicolumn{1}{c}{\textit{pure fine-tuned model}} \\ \hline
\multicolumn{3}{c}{{\textbf{lista esami primo anno di medicina}}} \\ \hline
\multicolumn{1}{l}{{\begin{tabular}[c]{@{}l@{}}
Gli esami del primo anno del corso di medicina\\
all'Università degli Studi di Palermo per l'anno\\
accademico 2023/24 non sono specificati nei\\
documenti forniti. Ti suggerisco di consultare i\\
bandi di concorso per maggiori informazioni.\\
Inoltre, per quanto riguarda l'immatricolazione,\\
i corsi di studio ad accesso libero possono\\
essere immatricolati senza mora dall'1 agosto\\
{[}...{]}\end{tabular}}} & \multicolumn{1}{l}{{\begin{tabular}[c]{@{}l@{}}
Mi dispiace, ma non ho trovato informazioni\\
riguardanti gli esami del primo anno del corso\\
di medicina all'Università degli Studi di\\
Palermo per l'anno accademico 2023/24. Ti\\
consiglio di consultare i bandi di concorso\\
per maggiori informazioni.\end{tabular}}} & \begin{tabular}[c]{@{}l@{}}
La lista degli esami del primo anno del corso\\
di laurea in Medicina e Chirurgia presso\\
l'Università di Palermo prevede i seguenti esami:\\
Anatomia Umana con Esercitazioni, Biologia Applicata,\\
Lingua Inglese, Chimica Biologica con Elementi di\\
Biologia Molecolare, Fisica, Matematica, Statistica e\\
Informatica, Istologia Umana con Elementi di Biologia\\
Cellulare, Fisiologia Umana con Esercitazioni, e\\
Patologia Generale.\end{tabular} \\ \hline
\multicolumn{3}{c}{{\textbf{quali sono le materie del primo anno di medicina}}} \\ \hline
\multicolumn{1}{l}{{\begin{tabular}[c]{@{}l@{}}
Le materie del primo anno del corso di medicina\\
all'Università degli Studi di Palermo per l'anno\\
accademico 2023/24 sono: Fisica I, Matematica I e\\
Chimica Generale ed Organica C.I. Invece, per il\\
corso di laurea in Scienze Biologiche, le materie\\
del primo anno sono: Matematica con esercitazioni,\\
Citologia e Istologia con esercitazioni, Chimica\\
Generale con esercitazioni e Botanica Generale e\\
istematica con esercitazioni.\end{tabular}}} & \multicolumn{1}{l}{{\begin{tabular}[c]{@{}l@{}}
Le materie del primo anno del corso di medicina\\
all'Università degli Studi di Palermo per l'anno\\
accademico 2023/24 sono: Fisica I, Matematica I,\\
Chimica Generale ed Organica C.I., Chimica\\
Organica, Biologia, Embriologia ed Istologia\\
C.I., Genetica e Microbiologia C.I., Informatica,\\
Bioinformatica e Inglese Tecnico Scientifico C.I.\\
Ti consiglio di consultare i documenti forniti\\
per maggiori informazioni.\end{tabular}}} & \begin{tabular}[c]{@{}l@{}}
Le materie del primo anno di medicina sono: Anatomia\\
Umana, Istologia, Embriologia, Biochimica, \\
Biologia Cellulare, Genetica, Fisica, Chimica, Lingua\\
Inglese, Metodologia del Lavoro Intellettuale, e\\
Cultura e Valori in Sanità.\end{tabular}\\ \hline
\end{tabular}}
\end{center}
\label{med-table}
\end{table}

\subsection{The SHARPER night}\label{sharper}

\begin{figure}[ht]
\includegraphics[width=\textwidth]{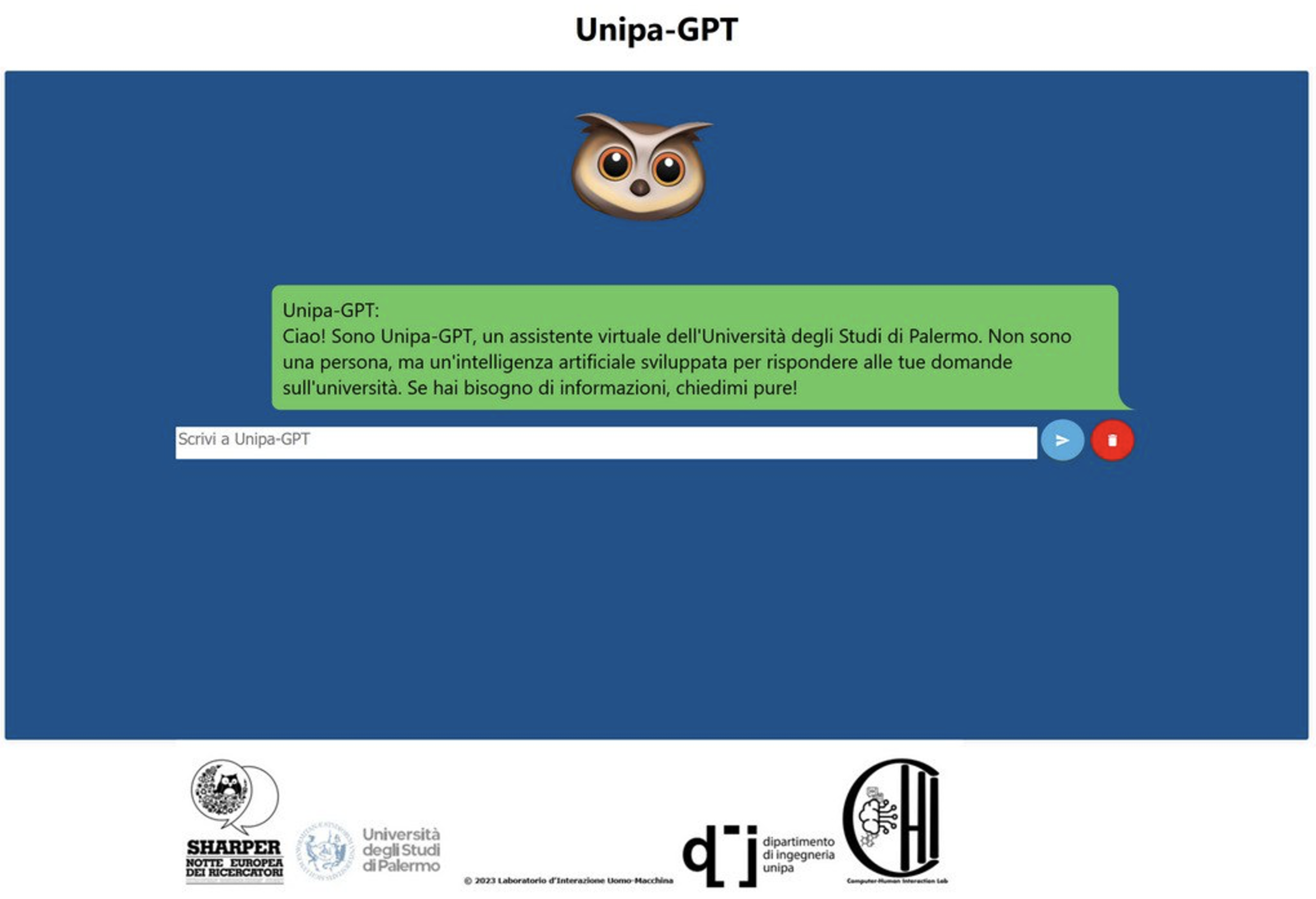}
\caption{The GUI of Unipa-GPT presented at the SHARPER night.}
\label{interface}
\end{figure}

Unipa-GPT was presented during the European Researchers’ Night (SHARPER night) when visitors were allowed to interact with the chat-bot making questions about the University of Palermo. The created interface, shown in Figure \ref{interface}, was designed to be as user-friendly as possible with an avatar that mimics the answer of the system by moving itself and reproducing an audio version of the answer, thus making it more accessible and interactive. As shown in \ref{interface} the chat-bot always starts a conversation displaying a welcome message in Italian, here reported in its English version: 

\begin{quote}
    Hi! I am Unipa-GPT, a virtual assistant at the University of Palermo. I am not a person, but an artificial intelligence developed to answer your questions about the university. If you need any information, just ask me!
\end{quote}

Some users were confident during the interaction, while others needed an external human-guidance to interact with the chat-bot, thus demonstrating that skeptical visitors about AI, in a real-context, are less prone to give a chance to this type of systems. The model that was actually used during the SHARPER Night is the one that uses \textit{condensed prompt} with \verb|gpt-3.5-turbo| and \verb|unipa-corpus-clear| as the retrieval corpus: due to further experiments, few changes were made in the prompt, as shown in Table \ref{prompt-sharper} (English version in Table \ref{prompt-sharper-en} in Appendix).

\begin{table}[ht!]
\caption{The prompts used in the system.}
\begin{center}
\resizebox{\columnwidth}{!}{
\begin{tabular}{ll}
\hline
\textit{prompt type}&\textit{prompt text} \\
\hline
\textit{custom prompt} & \begin{tabular}[c]{@{}l@{}}Sono Unipa-GPT, chatbot e assistente virtuale dell'Università degli Studi di Palermo\\  che risponde cordialmente e in forma colloquiale.\\
Ai saluti, rispondi salutando e presentandoti;\\
Rispondi alla domanda con la dicitura "Risposta: "\\
Ricordati che il rettore dell'Università è il professore Massimo Midiri.\\
Se la domanda riguarda l'università degli studi di Palermo,\\ rispondi in base alle informazioni e riporta i link ad esse associate;\\
Se non sai rispondere alla domanda, rispondi dicendo che sei un'intelligenza artificiale\\ che ha ancora molto da imparare e suggerisci\\ di andare su https://www.unipa.it/, non inventare risposte.\\
Domanda: \{question\}\\
Informazioni: \{context\}\end{tabular} \\
\hline
\textit{condensed prompt} & \begin{tabular}[c]{@{}l@{}}Data la seguente conversazione e la domanda successiva, riformula la domanda successiva \\ in modo tale sia una domanda singola.\\ Conversazione: \{chat\_history\}\\ Domanda succesiva: \{question\}\\ Domanda singola:\end{tabular} \\
\hline
\end{tabular}}
\end{center}
\label{prompt-sharper}
\end{table}

\begin{figure}[ht]
\includegraphics[width=\textwidth]{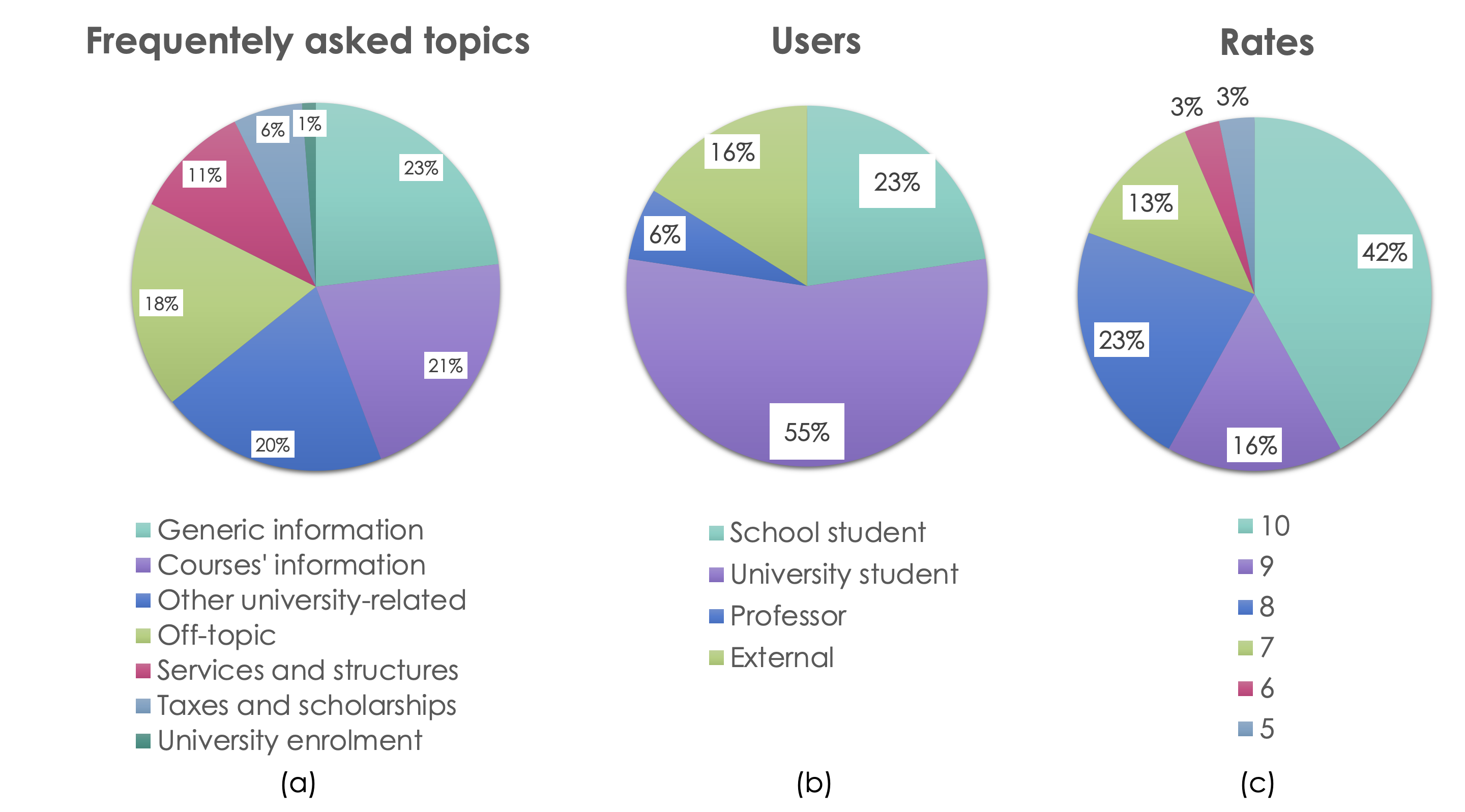}
\caption{Charts illustrate the frequently asked topics to the model during the SHARPER Night (a); among the 31 feedback collected, an overview of the users is reported (b) and the rates to the application (c).}
\label{charts-sharper}
\end{figure}

During the event, we collected a total of 165 questions that can be divided in seven categories, namely Generic Information, Courses' Information, Other University-related, Off-topic, Services and Structures, Taxes and Scholarships and University Environment (Figure \ref{charts-sharper}.a). On average, the interaction produced a two questions long chat, and we asked each visitor to give a feedback via a Google Form thus collecting 31 replies. The majority of the collected feedback came from either University students or professors, i.e. people that are familiar with the academic world (Figure \ref{charts-sharper}.b) and the general rate of the chat-bot is very positive, despite two bad rankings (Figure \ref{charts-sharper}.c). One bad evaluation came from a conversation where a student asked about a specific parameter in the scholarship rank: in this particular case the model provided a false answer, saying that the asked parameter is not considered for the rank, despite the truth. We suppose that this is a case in which the model made use of an incomplete document. The second bad evaluation came from a conversation where information about the final test for a bachelor course were asked to the model. Here the problem is the bachelor course name: the user did not used the complete official course name but the one to which it is usually referred to. Moreover, bachelor and master courses share the same name, and this may cause an ambiguity in the model that can be fixed by adding external information. In this case, the final test is peculiar of the bachelor course while thesis refers to the master course: we assume that if this kind of general knowledge is provided to the model, in addition to the document, the correct answer may be generated. 

\subsection{gpt-3.5-turbo vs other open LLMs}\label{second}

During the first semester of 2024 we carried out more experiments to evaluate the quality of the answers provided by Unipa-GPT, and we compared the performances of \verb|gpt-3.5-turbo|, considered as a baseline model, with other LLMs averaging from 3B to 8B parameters, considering both foundation and fine-tuned models. The choice of the models sizes relies only on computational constraints as we did not have access to any cloud with HPC facility at that time, and we used local resources.

We selected one question for each typology reported in Figure \ref{charts-sharper}.a) picking them from the real ones posed by the users to Unipa-GPT during the SHARPER Night, and excluding the off-topic requests. We built manually a golden answer to each question as it is shown in Table \ref{goldenqa} (Table \ref{goldenqa-en} in Appendix for the English version).

\begin{table}[ht!]
\caption{List of  the six QA pairs used for evaluation purposes}
\begin{center}
\resizebox{\columnwidth}{!}{
\begin{tabular}{cll}
\hline
\textbf{ID}&\textbf{Question} & \textbf{Answer} \\ \hline
\textbf{Q1}&
\begin{tabular}[c]{@{}l@{}}
\textit{Chi è il professore di Intelligenza}\\ 
\textit{Artificiale 1 per il corso di}\\ 
\textit{Laurea Magistrale in Ingegneria}\\ 
\textit{Informatica?}\end{tabular} &
  \begin{tabular}[c]{@{}l@{}}
Il professore di Intelligenza Artificiale 1 del corso di Laurea Magistrale\\
Magistrale in Ingegneria Informatica è il professore Gaglio e verrà\\
erogata durante il primo semestre. Per maggiori informazioni vai su\\
http://www.unipa.it/struttura.html?id=721\end{tabular} \\ \hline
\textbf{Q2}&
\begin{tabular}[c]{@{}l@{}}
\textit{Quali sono le scadenze di}\\ 
\textit{iscrizione ad un corso di laurea?}\\ \end{tabular} &
  \begin{tabular}[c]{@{}l@{}}
La presentazione delle domande di iscrizione per l'Anno Accademico\\
2023/2024 varia in base alla tipologia di corso. Per i corsi di studio\\
triennali e magistrali a ciclo unico, la domanda può essere presentata\\ dall'1 agosto al 30 settembre 2023, mentre per i corsi magistrali,\\
dall'1 agosto al 30 novembre 2023, in entrambi i casi, è richiesto il\\
pagamento della prima rata delle tasse universitarie. Per i corsi di\\
studio ad accesso programmato locale, con prenotazione online o ad\\
accesso programmato nazionale, le immatricolazioni vanno effettuate\\
entro i termini previsti dal relativi bandi di concorso.\end{tabular} \\ \hline
\textbf{Q3}&
\begin{tabular}[c]{@{}l@{}}
\textit{Come funziona la magistrale}\\  
\textit{di chimica?}\end{tabular} &
  {\begin{tabular}[c]{@{}l@{}}
Il corso di laurea magistrale in Chimica presso l'Università degli Studi\\
di Palermo, ha sede a Palermo e ha una durata di 2 anni. L'obiettivo del\\
corso è la formazione di laureati con una solida preparazione di base,\\
che li ponga in grado di affrontare con competenza ed autonomia sia\\
attività professionali sia quelle di ricerca accademica ed industriale.\\ 
Per questo motivo il corso di studio è progettato con lo scopo di\\
completare la cultura scientifica nelle discipline chimiche fondamentali\\
e di introdurre conoscenze più avanzate. Quindi lo studente può\\
completare il proprio piano di studi, attenzionando gli insegnamenti\\
che meglio si adattano ai propri interessi specifici,  come la chimica dei\\
materiali, dell'ambiente e dei beni culturali, la chimica supramolecolare\\
e la didattica della chimica. Per maggiori informazioni vai su\\ 
http://www.unipa.it/struttura.html?id=766\end{tabular}} \\ \hline
\textbf{Q4}&
\begin{tabular}[c]{@{}l@{}}
\textit{ciao! sono un ragazzo appena uscito}\\ 
\textit{dal liceo che è interessato al settore}\\ 
\textit{legale,in particolare alle leggi sulle}\\ 
\textit{aziende. dove potrei iscrivermi?}\end{tabular} &
  \begin{tabular}[c]{@{}l@{}}

Ciao! Il corso di Laurea in Consulente Giuridico d'Impresa sembra\\
proprio fare al caso tuo! Il corso di laurea ha sede a Trapani e ha\\
una durata di 3 anni e ti fornirà le conoscenze in ambito giuridico ed\\ economico-aziendalistico. Una volta terminato il percorso di studio\\
potrai svolgere attività interdisciplinari che richiedono competenze\\ giuridiche, aziendalistiche e organizzativo-gestionali. Per maggiori\\ 
informazioni vai su http://www.unipa.it/struttura.html?id=1557\end{tabular} \\ \hline
\textbf{Q5}&
\begin{tabular}[c]{@{}l@{}}
\textit{come posso prenotare un}\\ 
\textit{appuntamento in segreteria?}\end{tabular} &
    \begin{tabular}[c]{@{}l@{}}
È possibile recarsi in segreteria il lunedì, mercoledì e venerdì dalle\\
10.00 alle 12.00, martedì e giovedì dalle 15.00 alle 17.00. Puoi\\
prenotare il tuo turno attraverso la App SolariQ. Per maggiori\\
informazioni vai su\\
https://www.unipa.it/servizi/segreterie/\end{tabular} \\ \hline
\textbf{Q6}&
\textit{Come si pagano le tasse?} &
  \begin{tabular}[c]{@{}l@{}}

Il pagamento delle tasse deve essere effettuato esclusivamente\\
mediante sistema PAgoPA (Pagamenti della Pubblica Amministrazione)\\
al quale si accede tramite portale di ateneo. Dopo aver compilato la\\ pratica online, è possibile pagare direttamente online con il sistema\\ PAgoPA o stampare il bollettino e pagare presso una ricevitoria\\
abilitata PAgoPA. Ulteriori informazioni sul pagamento via PAgoPA\\
sono reperibili qui https://immaweb.unipa.it/immaweb/public/\\
pagamenti.seam, mentre è disponibile il Regolamento in materia di\\ contribuzione studentesca https://www.unipa.it/servizi/\\
segreterie/.content/documenti/regolamenti\_calendari/2023/\\
5105144-def\_regolamento-contribuzione--studentesca\\
-2023---24-2.pdf\end{tabular} \\ \hline
\end{tabular}}
\end{center}
\label{goldenqa}
\end{table}

The following foundation models were selected: \texttt{Llama-2} \cite{touvron2023llama}, \texttt{LLaMAntino-2} \cite{basile2023llamantino}, \texttt{Llama-3} \cite{llama3}, \texttt{Minerva} \cite{minerva}.

We considered both chat-oriented and purely fine-tuned models: \texttt{Llama-2-chat} \cite{touvron2023llama}, \texttt{LLaMAntino-2-chat} \cite{basile2023llamantino}, \texttt{Fauno} \cite{bacciu2023fauno}, and instruction-tuned models: \texttt{Camoscio} \cite{santilli2023camoscio}, \texttt{Llama-3-instruct} \cite{llama3}, and \texttt{LLaMAntino-3-ANITA} \cite{polignano2024advanced} henceforth referred to as \verb|Anita-3|.

All models, except for \texttt{Minerva}, that is a \verb|Mistral|-based model \cite{jiang2023mistral7b}, are \verb|Llama|-based ones. In particular, \texttt{LLaMAntino-2}, \texttt{Fauno} and \texttt{Camoscio} are fine-tuned version of \texttt{Llama-2}, while \verb|LLaMAntino-2-chat| is a build upon \verb|Llama-2-chat|, and \verb|Anita-3| is a fine-tuned version of \verb|Llama-3-instruct|. 

Evaluation was performed through instruction prompting using the \emph{custom prompt}, and models were forced to generate a maximum of 256 new tokens. Except for \verb|gpt-3.5-turbo|, all the evaluation runs were performed on a 48 GB NVIDIA RTX 6000 Ada Generation without any quantization. Even if the same prompt was used for all the models, it was provided differently to each of them following its authors' specific prompt strategies for either fine-tuning or inference. As an example, \verb|### Istruzione:| was added before the prompt for \verb|Camoscio|, \verb|Minerva-3B|, \verb|LLaMAntino-2|, \verb|Llama-3| and \verb|Llama-2|, while the \verb|[/INST]| tag was used for \verb|Llama-2-chat| . Moreover, exactly the same retriever was used in all the runs in order to provide each model with the same documents of the othr ones, thus measuring its capability to paraphrase, and re-elaborate the provided context for generating a helpful answer. 

\begin{table}[ht!]
\caption{Results with BLEU score. Bold values are the higher ones.}
\begin{center}
\begin{tabular}{lcccccc}
\hline
&Q1&Q2&Q3&Q4&Q5&Q6\\
\hline
\multicolumn{1}{c}{\textit{Baseline models}}&&&&&&\\
\hline
gpt-3.5-turbo&\textbf{0.334}&\textbf{0.151}&\textbf{0.309}&\textbf{0.213}&\textbf{0.207}&\textbf{0.197}\\
\hline
\multicolumn{1}{c}{\textit{Foundation models}}&&&&&&\\
\hline
Llama-2&0.0&0.0&0.0&0.0&0.0&0.0\\
LLaMantino-2&0.0&0.0&0.0&0.0&0.0&0.0\\
Llama-3&0.0&0.0&0.0&0.0&0.0&0.0\\
Minerva-3B&0.0&0.0&0.0&0.0&0.0&0.0\\
\hline
\multicolumn{1}{c}{\textit{Chat-oriented fine-tuned models}}&&&&&&\\
\hline
Llama-2-chat&0.0&0.0&0.0&0.0&0.0&0.0\\
LLaMantino-2-chat&0.0&0.0&0.0&0.0&0.0&0.0\\
Fauno&0.0&0.0&0.0&0.0&0.0&0.0\\
\hline
\multicolumn{1}{c}{\textit{Instruction-tuned models}}&&&&&&\\
\hline
Camoscio&0.0&0.0&0.0&0.0&0.019&0.0\\
Llama-3-instruct&0.047&0.0&0.114&0.0&0.091&0.066\\
Anita-3&0.048&0.0&0.066&0.063&0.0&0.068\\
\hline
\end{tabular}
\end{center}
\label{bleu}
\end{table}



\begin{table}[ht!]
\caption{Results with ROUGE-L score. Bold values are the higher ones.}
\begin{center}
\begin{tabular}{lcccccc}
\hline
&Q1&Q2&Q3&Q4&Q5&Q6\\
\hline
\multicolumn{1}{c}{\textit{Baseline models}}&&&&&&\\
\hline
gpt-3.5-turbo&\textbf{0.472}&\textbf{0.295}&\textbf{0.423}&\textbf{0.301}&\textbf{0.187}&\textbf{0.284}\\
\hline
\multicolumn{1}{c}{\textit{Foundation models}}&&&&&&\\
\hline
Llama-2&0.054&0.049&0.007&0.010&0.020&0.008\\
LLaMantino-2&0.035&0.042&0.050&0.014&0.045&0.044\\
Llama-3&0.052&0.012&0.020&0.051&0.020&0.009\\
Minerva-3B&0.023&0.030&0.0&0.0&0.054&0.043\\
\hline
\multicolumn{1}{c}{\textit{Chat-oriented fine-tuned models}}&&&&&&\\
\hline
Llama-2-chat&0.010&0.009&0.013&0.009&0.008&0.019\\
LLaMantino-2-chat&0.0&0.05&0.039&0.0&0.0&0.0\\
Fauno&0.013&0.046&0.013&0.011&0.033&0.047\\
\hline
\multicolumn{1}{c}{\textit{Instruction-tuned models}}&&&&&&\\
\hline
Camoscio&0.010&0.016&0.019&0.019&0.031&0.017\\
Llama-3-instruct&0.109&0.052&0.131&0.091&0.092&0.074\\
Anita-3&0.133&0.124&0.118&0.078&0.096&0.089\\

\hline
\end{tabular}
\end{center}
\label{rouge}
\end{table}

Correctness of the answers with respect to the golden labels was measured both with the  BLEU~\cite{papineni2002bleu} and the ROGUE-L score~\cite{lin2004rouge}. It is worth noticing that BLEU and ROUGE are not specifically tailored for evaluating a text generation task, but they can be considered as a good starting point in automatic evaluation, given a golden answer. 

The results are reported in Table \ref{bleu} and Table \ref{rouge}, and they show clearly that  \verb|gpt-3.5-turbo| performs significantly better than both foundation and chat-oriented fine-tuned models that attain 0. Only instruction-tuned models obtain a non-zero BLEU score, but such values remain far low compared with the results for \verb|gpt-3.5-turbo|.

Results in Table \ref{rouge} for the ROUGE-L score confirm the aforementioned comments for BLEU: \verb|gpt-3.5-turbo| performs significantly better compared to other models, and instruction-tuned models reach the highest results in comparison with the other \verb|Llama|- and \verb|Mistral|-based ones. 

We also assessed the answers provided by each model through manual inspection. A tendency comes up to generate correct English answers (\verb|Llama-2-chat|), out of topic answers bot in Italian (\verb|LLaMantino-2| models and \verb|Minerva|) and other languages (\verb|Llama-2| and \verb|Camoscio|) or including code (\verb|Fauno| and \verb|Llama-3|). \verb|Llama-3-instruct| and \verb|Anita-3| generate the most satisfactory answers, since they provide topic-related answers in Italian apart from few English ones.

Further evaluations were carried out using RAGAS to compute \textit{context relevancy}, \textit{faithfulness}, and \textit{answer correctness}. The RAGAS metrics need a ``judge'', and we opted for \verb|gpt-4-turbo| as is the most performative model we had access to at the time of the experiments.

Context relevancy is measured based on both the question itself and the retrieved documents that form the context. The values fall within the range of (0, 1) with higher values indicating better relevancy. Results are reported in Table \ref{ragas-context}, where the scores in each row refer to the first four documents retrieved by the RAG to answer the corresponding question. We starred the values obtained for the ``golden documents'',  that is the ones used for generating each of the golden answers, in order to highlight divergences. The results are extremely low, and a correct retrieval (i.e. the document with highest context relevancy coincides with a golden document) is verified in just three out of the six  considered questions. We argue that this finding could be caused by the huge documents' size compared with the length of the question, and the actual relevant piece of information.

\begin{table}[ht!]
\caption{Evaluation results with RAGAS context relevancy metric. Starred results refer to the document that was used for manual generation of the respective golden answer, while bold values refer to the highest scores.}
\begin{center}
\begin{tabular}{ccccc}
\hline
& D1 & D2 & D3 & D4 \\
\hline
Q1 & \textbf{*0.1*} & \textbf{0.1} & 0.083 & 0.071 \\
Q2 & *0.091* & 0.125 & \textbf{*0.625*} & 0.333 \\
Q3 & *0.1* & \textbf{0.111} & \textbf{0.111} & \textbf{0.111} \\
Q4 & \textbf{*0.167*} & *0.059* & 0.1 & 0.333 \\
Q5 & *0.429* & \textbf{0.5} & 0.143 & 0.111 \\
Q6 & \textbf{1.0} & *0.1* & 0.167 & 0.5 \\
\hline
\end{tabular}
\end{center}
\label{ragas-context}
\end{table}

\begin{table}[ht!]
\caption{Evaluation results with RAGAS faithfulness metric. Bold values are the highest ones.}
\begin{center}
\begin{tabular}{lcccccc}
\hline
&Q1&Q2&Q3&Q4&Q5&Q6\\
\hline
\multicolumn{1}{c}{\textit{Baseline models}}&&&&&&\\
\hline
gpt-3.5-turbo&\textbf{1.0}&\textbf{1.0}&\textbf{1.0}&\textbf{1.0}&0.333&\textbf{0.875}\\
\hline
\multicolumn{1}{c}{\textit{Foundation models}}&&&&&&\\
\hline
Llama-2&0.083&0.0&0.0&0.0&0.0&0.0\\
LLaMantino-2&0.0&0.0&0.0&0.0&0.0&0.0\\
Llama-3&0.2&0.5&0.0&0.0&0.0&0.0\\
Minerva-3B&0.0&0.0&0.0&0.0&0.0&0.0\\
\hline
\multicolumn{1}{c}{\textit{Chat-oriented fine-tuned models}}&&&&&&\\
\hline
Llama-2-chat&0.0&0.2&0.0&0.0&0.0&0.0\\
LLaMantino-2-chat&0.0&0.0&0.0&0.0&0.0&0.0\\
Fauno&0.0&0.0&0.0&0.0&0.25&0.286\\
\hline
\multicolumn{1}{c}{\textit{Instruction-tuned models}}&&&&&&\\
\hline
Camoscio&0.0&0.0&0.0&0.0&0.0&0.0\\
Llama-3-instruct&0.143&\textbf{1.0}&0.25&0.0&0.667&0.0\\
Anita-3&0.429&0.6&0.667&0.444&\textbf{1.0}&0.333\\

\hline
\end{tabular}
\end{center}
\label{ragas-faith}
\end{table}

\begin{table}[ht!]
\caption{Evaluation results with RAGAS answer correctness metric. Bold values are the highest ones.}
\begin{center}
\begin{tabular}{lcccccc}
\hline
&Q1&Q2&Q3&Q4&Q5&Q6\\
\hline
\multicolumn{1}{c}{\textit{Baseline models}}&&&&&&\\
\hline
gpt-3.5-turbo&\textbf{0.616}&\textbf{0.727}&\textbf{0.563}&0.401&\textbf{0.531}&\textbf{0.666}\\
\hline
\multicolumn{1}{c}{\textit{Foundation models}}&&&&&&\\
\hline
Llama-2&0.21&0.183&0.175&0.173&0.18&0.29\\
LLaMantino-2&0.191&0.186&0.18&0.188&0.186&0.187\\
Llama-3&0.385&0.429&0.199&0.199&0.191&0.184\\
Minerva-3B&0.183&0.186&0.176&0.189&0.195&0.203\\
\hline
\multicolumn{1}{c}{\textit{Chat-oriented fine-tuned models}}&&&&&&\\
\hline
Llama-2-chat&0.175&0.314&0.179&0.196&0.19&0.216\\
LLaMantino-2-chat&0.187&0.184&0.182&0.188&0.188&0.191\\
Fauno&0.175&0.213&0.152&0.162&0.182&0.219\\
\hline
\multicolumn{1}{c}{\textit{Instruction-tuned models}}&&&&&&\\
\hline
Camoscio&0.287&0.175&0.173&0.171&0.18&0.181\\
Llama-3-instruct&0.396&0.22&0.322&0.217&0.224&0.605\\
Anita-3&0.395&0.362&0.541&\textbf{0.444}&0.515&0.224\\

\hline
\end{tabular}
\end{center}
\label{ragas-correct}
\end{table}



The faithfulness measures the factual consistency of the generated answer against the given context, while answer correctness involves gauging the accuracy of the generated answer when compared to the ground truth. Both metrics range from 0 to 1 and better performances are associated with higher scores. The results are reported in Table \ref{ragas-faith} and \ref{ragas-correct} for the faithfulness and the answer correctness, respectively.

The highest faithfulness results are reached by \verb|gpt-3.5-turbo|, followed by \verb|Llama-3-instruct|, and \verb|Anita-3|, confirming the previous results for both BLEU and ROUGE. The two models performed far better than \verb|gpt-3.5-turbo| in Q5. Similar behavior can be found in answer correctness, where \verb|Anita-3| outperformed \verb|gpt-3.5-turbo| in Q4, while \verb|Llama-3-instruct| received the same score as \verb|gpt-3.5-turbo| in Q2. After manual inspection we found that the answers generated by \verb|Anita-3| convey the correct meaning, but they are not well spelled, and are grammatically wrong, thus being penalized both with traditional metrics and with RAGAS ones.

The presented results show clearly that \verb|gpt-3.5-turbo| was a SOTA model for the time Unipa-GPT was released, and in general the GPT family is an ideal one for the kind of architecture we presented in this work. We can not exclude that 13B to 70B models built on both the \verb|Llama| and the \verb|Mistral| foundation ones using either intstuction tuning or explicit fine-tuning could reach comparable performances.

\section{Conclusions and future works}\label{conc}

In this paper we presented Unipa-GPT, a virtual assistant capable of answering to secondary school students who need information for accessing and studying at the University of Palermo. The developed system relies on a RAG architecture that uses documents from a corpus purposely scraped from the University institutional website and comes in two versions that make use of either \verb|gpt-3.5-turbo| or a fine-tuned model \verb|gpt-3.5-turbo-fine-tuned| where the corpus has been reduced to keep low the computational resources needed for fine-tuning. Moreover, different versions of the system have been tested as regards both the prompt and the strutcure of the documents used in the RAG corpus. Significant improvements were not found in the fine-tuned model, and the best performing system was the one that uses the so called \emph{condensed prompt} where the previous conversation and the current question are reformulated into a unique prompt. Such a prompt induces \verb|gpt-3.5-turbo| to summarize the conversation at each question, and then it behaves as it was instructed instructed using our base \emph{custom prompt} which is tailored for the application purposes. Moreover, this system uses the \verb|unipa-corpus-clear| for the retrieval where the specifico educational objectives for each class are not reported. We argue that this ``light'' version of the corpus provides more compact and precise information to the LLM thus generating the best answers. The presented model has been compared against several open LLMs based on both the \verb|Llama| and the \verb|Mistral| foundation ones using different metrics, and the results proved that the \verb|gpt-| family outperforms the others in this task.
At the time of writing the manuscript, we moved to the integration of \verb|gpt-4.o-| models in the Unipa-GPT framework. Future work will be devoted to build an open framework where all kinds of models can be integrated and tested. Finally, it is well known that integrating structured knowledge in RAG can mitigate hallucination \cite{CASHEEKAR2024100632}. Following this assumption, we are developing a hybrid RAG for the Unipa-GPT framework, that relies on both a vector store, and either a knowledge graph obtained by document indexing or a formal OWL ontology if available to describe the domain under investigation. 

\begin{acknowledgments}
  We thank all the CHILab team, in particular Dr. Salvatore Contino and  Dr. Paolo Sortino, and PhD students Luca Cruciata and Gaetano Pottino, that contributed in generating the corpus with the scraping and implemented the demonstration interface.
  
\end{acknowledgments}

\bibliographystyle{fullname}
\bibliography{bibliography}

\begin{thebibliography}{}

\bibitem[\protect\citename{Bacciu \bgroup et al.\egroup }2023]{bacciu2023fauno}
Bacciu, Andrea, Giovanni Trappolini, Andrea Santilli, Emanuele Rodol{\`a}, and Fabrizio Silvestri.
\newblock 2023.
\newblock Fauno: The italian large language model that will leave you senza parole!
\newblock {\em arXiv preprint arXiv:2306.14457}.

\bibitem[\protect\citename{Basile \bgroup et al.\egroup }2023]{basile2023llamantino}
Basile, Pierpaolo, Elio Musacchio, Marco Polignano, Lucia Siciliani, Giuseppe Fiameni, and Giovanni Semeraro.
\newblock 2023.
\newblock Llamantino: Llama 2 models for effective text generation in italian language.
\newblock {\em arXiv preprint arXiv:2312.09993}.

\bibitem[\protect\citename{Bhat \bgroup et al.\egroup }2024]{restaurantchat}
Bhat, Vani, Sree~D. Cheerla, Jinu~R. Mathew, Nupur Pathak, Guannan Liu, and Jerry Gao.
\newblock 2024.
\newblock Retrieval augmented generation (rag) based restaurant chatbot with ai testability.
\newblock In {\em 2024 IEEE 10th International Conference on Big Data Computing Service and Machine Learning Applications (BigDataService)}, pages 1--10, Shanghai, China, July.

\bibitem[\protect\citename{Borji}2023]{borji2023categorical}
Borji, Ali.
\newblock 2023.
\newblock A categorical archive of chatgpt failures.
\newblock {\em arXiv preprint arXiv:2302.03494}.

\bibitem[\protect\citename{Brown \bgroup et al.\egroup }2020]{brown2020language}
Brown, Tom, Benjamin Mann, Nick Ryder, Melanie Subbiah, Jared~D. Kaplan, Prafulla Dhariwal, Arvind Neelakantan, Pranav Shyam, Girish Sastry, Amanda Askell, et~al.
\newblock 2020.
\newblock Language models are few-shot learners.
\newblock {\em Advances in neural information processing systems}, 33:1877--1901.

\bibitem[\protect\citename{Casheekar \bgroup et al.\egroup }2024]{CASHEEKAR2024100632}
Casheekar, Avyay, Archit Lahiri, Kanishk Rath, Kaushik~S. Prabhakar, and Kathiravan Srinivasan.
\newblock 2024.
\newblock A contemporary review on chatbots, ai-powered virtual conversational agents, chatgpt: Applications, open challenges and future research directions.
\newblock {\em Computer Science Review}, 52:100632.

\bibitem[\protect\citename{Devlin \bgroup et al.\egroup }2019]{devlin-etal-2019-bert}
Devlin, Jacob, Ming-Wei Chang, Kenton Lee, and Kristina Toutanova.
\newblock 2019.
\newblock {BERT}: Pre-training of deep bidirectional transformers for language understanding.
\newblock In Jill Burstein, Christy Doran, and Thamar Solorio, editors, {\em Proceedings of the 2019 Conference of the North {A}merican Chapter of the Association for Computational Linguistics: Human Language Technologies, Volume 1 (Long and Short Papers)}, pages 4171--4186, Minneapolis, Minnesota, June. Association for Computational Linguistics.

\bibitem[\protect\citename{Dwivedi \bgroup et al.\egroup }2023]{dwivedi2023so}
Dwivedi, Yogesh~K., Nir Kshetri, Laurie Hughes, Emma~L. Slade, Anand Jeyaraj, Arpan~K. Kar, Abdullah~M. Baabdullah, Alex Koohang, Vishnupriya Raghavan, Manju Ahuja, et~al.
\newblock 2023.
\newblock “so what if chatgpt wrote it?” multidisciplinary perspectives on opportunities, challenges and implications of generative conversational ai for research, practice and policy.
\newblock {\em International Journal of Information Management}, 71:102642.

\bibitem[\protect\citename{Es \bgroup et al.\egroup }2024]{es2023ragas}
Es, Shahul, Jithin James, Luis Espinosa~Anke, and Steven Schockaert.
\newblock 2024.
\newblock {RAGA}s: Automated evaluation of retrieval augmented generation.
\newblock In Nikolaos Aletras and Orphee De~Clercq, editors, {\em Proceedings of the 18th Conference of the European Chapter of the Association for Computational Linguistics: System Demonstrations}, pages 150--158, St. Julians, Malta, March. Association for Computational Linguistics.

\bibitem[\protect\citename{Forootani, Aliabadi, and Thraen}2024]{forootani2024bioenglmmaiassistchatbot}
Forootani, Ali, Danial~E. Aliabadi, and Daniela Thraen.
\newblock 2024.
\newblock Bio-eng-lmm ai assist chatbot: A comprehensive tool for research and education.
\newblock {\em arXiv preprint arXiv:2409.07110}.

\bibitem[\protect\citename{Gill \bgroup et al.\egroup }2024]{GILL202419}
Gill, Sukhpal~S., Minxian Xu, Panos Patros, Huaming Wu, Rupinder Kaur, Kamalpreet Kaur, Stephanie Fuller, Manmeet Singh, Priyansh Arora, Ajith~K. Parlikad, et~al.
\newblock 2024.
\newblock Transformative effects of chatgpt on modern education: Emerging era of ai chatbots.
\newblock {\em Internet of Things and Cyber-Physical Systems}, 4:19--23.

\bibitem[\protect\citename{Holtzman \bgroup et al.\egroup }2020]{holtzman2019curious}
Holtzman, Ari, Jan Buys, Li~Du, Maxwell Forbes, and Yejin Choi.
\newblock 2020.
\newblock The curious case of neural text degeneration.
\newblock In {\em International Conference on Learning Representations}, Addis Ababa, Ethiopia, April.

\bibitem[\protect\citename{Hu \bgroup et al.\egroup }2022]{hu2021lora}
Hu, Edward~J., Yelong Shen, Phillip Wallis, Zeyuan Allen-Zhu, Yuanzhi Li, Shean Wang, Lu~Wang, and Weizhu Chen.
\newblock 2022.
\newblock Lo{RA}: Low-rank adaptation of large language models.
\newblock In {\em International Conference on Learning Representations}, Online, April.

\bibitem[\protect\citename{Ji \bgroup et al.\egroup }2023]{10.1145/3571730}
Ji, Ziwei, Nayeon Lee, Rita Frieske, Tiezheng Yu, Dan Su, Yan Xu, Etsuko Ishii, Ye~Jin Bang, Andrea Madotto, and Pascale Fung.
\newblock 2023.
\newblock Survey of hallucination in natural language generation.
\newblock {\em ACM Computing Survey}, 55(12), mar.

\bibitem[\protect\citename{Jiang \bgroup et al.\egroup }2023]{jiang2023mistral7b}
Jiang, Albert~Q., Alexandre Sablayrolles, Arthur Mensch, Chris Bamford, Devendra~S. Chaplot, Diego de~las Casas, Florian Bressand, Gianna Lengyel, Guillaume Lample, Lucile Saulnier, et~al.
\newblock 2023.
\newblock Mistral 7b.
\newblock {\em arXiv preprint arXiv:2310.06825}.

\bibitem[\protect\citename{Johnson, Douze, and J{\'e}gou}2019]{johnson2019billion}
Johnson, Jeff, Matthijs Douze, and Herv{\'e} J{\'e}gou.
\newblock 2019.
\newblock Billion-scale similarity search with {GPUs}.
\newblock {\em IEEE Transactions on Big Data}, 7(3):535--547.

\bibitem[\protect\citename{Lewis \bgroup et al.\egroup }2020]{lewis2020retrieval}
Lewis, Patrick, Ethan Perez, Aleksandra Piktus, Fabio Petroni, Vladimir Karpukhin, Naman Goyal, Heinrich K{\"u}ttler, Mike Lewis, Wen-tau Yih, Tim Rockt{\"a}schel, et~al.
\newblock 2020.
\newblock Retrieval-augmented generation for knowledge-intensive nlp tasks.
\newblock {\em Advances in Neural Information Processing Systems}, 33:9459--9474.

\bibitem[\protect\citename{Lin}2004]{lin2004rouge}
Lin, Chin-Yew.
\newblock 2004.
\newblock {ROUGE}: A package for automatic evaluation of summaries.
\newblock In {\em Text Summarization Branches Out}, pages 74--81, Barcelona, Spain, July. Association for Computational Linguistics.

\bibitem[\protect\citename{Llama~Team}2024]{llama3}
Llama~Team, AI @~Meta.
\newblock 2024.
\newblock The llama 3 herd of models.
\newblock {\em arXiv preprint arXiv:2407.21783}.

\bibitem[\protect\citename{Lo}2023]{educsci13040410}
Lo, Chung~K.
\newblock 2023.
\newblock What is the impact of chatgpt on education? a rapid review of the literature.
\newblock {\em Education Sciences}, 13(4).

\bibitem[\protect\citename{OpenAI}2024]{openai2024gpt4technicalreport}
OpenAI.
\newblock 2024.
\newblock Gpt-4 technical report.
\newblock {\em arXiv preprint arXiv:2303.08774}.

\bibitem[\protect\citename{Orlando \bgroup et al.\egroup }2024]{minerva}
Orlando, Riccardo, Luca Moroni, Pere-Llu{\'\i}s~H. Cabot, Edoardo Barba, Simone Conia, Sergio Orlandini, Giuseppe Fiameni, and Roberto Navigli.
\newblock 2024.
\newblock Minerva llms: The first family of large language models trained from scratch on italian data.
\newblock {\em Proceedings of the 10th Italian Conference on Computational Linguistics (CLiC-it 2024), Pisa, Italy}, December.

\bibitem[\protect\citename{Papineni \bgroup et al.\egroup }2002]{papineni2002bleu}
Papineni, Kishore, Salim Roukos, Todd Ward, and Wei-Jing Zhu.
\newblock 2002.
\newblock Bleu: a method for automatic evaluation of machine translation.
\newblock In {\em Proceedings of the 40th annual meeting of the Association for Computational Linguistics}, pages 311--318, Philadelphia, Pennsylvania, USA, July.

\bibitem[\protect\citename{Polignano, Basile, and Semeraro}2024]{polignano2024advanced}
Polignano, Marco, Pierpaolo Basile, and Giovanni Semeraro.
\newblock 2024.
\newblock Advanced natural-based interaction for the italian language: Llamantino-3-anita.
\newblock {\em arXiv preprint arXiv:2405.07101}.

\bibitem[\protect\citename{Rahman and Watanobe}2023]{app13095783}
Rahman, Md.~Mostafizer and Yutaka Watanobe.
\newblock 2023.
\newblock Chatgpt for education and research: Opportunities, threats, and strategies.
\newblock {\em Applied Sciences}, 13(9).

\bibitem[\protect\citename{Rodriguez-Torrealba, Garcia-Lopez, and Garcia-Cabot}2022]{RODRIGUEZTORREALBA2022118258}
Rodriguez-Torrealba, Ricardo, Eva Garcia-Lopez, and Antonio Garcia-Cabot.
\newblock 2022.
\newblock End-to-end generation of multiple-choice questions using text-to-text transfer transformer models.
\newblock {\em Expert Systems with Applications}, 208:118258.

\bibitem[\protect\citename{Santilli and Rodolà}2023]{santilli2023camoscio}
Santilli, Andrea and Emanuele Rodolà.
\newblock 2023.
\newblock Camoscio: an italian instruction-tuned llama.
\newblock {\em arXiv preprint arXiv:2307.16456}.

\bibitem[\protect\citename{Sharma and Yadav}2023]{Sharma_Yadav_2023}
Sharma, Sudhansh and Ramesh Yadav.
\newblock 2023.
\newblock Chat gpt – a technological remedy or challenge for education system.
\newblock {\em Global Journal of Enterprise Information System}, 14(4):46--51, May.

\bibitem[\protect\citename{Tack and Piech}2022]{tack2022aiteachertestmeasuring}
Tack, Anaïs and Chris Piech.
\newblock 2022.
\newblock The {AI} teacher test: Measuring the pedagogical ability of blender and {GPT-3} in educational dialogues.
\newblock In Antonija Mitrovic and Nigel Bosch, editors, {\em Proceedings of the 15th International Conference on Educational Data Mining}, pages 522--529, Durham, United Kingdom, July. International Educational Data Mining Society.

\bibitem[\protect\citename{Touvron \bgroup et al.\egroup }2023]{touvron2023llama}
Touvron, Hugo, Louis Martin, Kevin Stone, Peter Albert, Amjad Almahairi, Yasmine Babaei, Nikolay Bashlykov, Soumya Batra, Prajjwal Bhargava, Shruti Bhosale, et~al.
\newblock 2023.
\newblock Llama 2: Open foundation and fine-tuned chat models.
\newblock {\em arXiv preprint arXiv:2307.09288}.

\bibitem[\protect\citename{Vakayil \bgroup et al.\egroup }2024]{vakayil2024rag}
Vakayil, Sonia, D.~Sujitha Juliet, Anitha J., and Sunil Vakayil.
\newblock 2024.
\newblock Rag-based llm chatbot using llama-2.
\newblock In {\em 2024 7th International Conference on Devices, Circuits and Systems (ICDCS)}, pages 1--5, Coimbatore, India, April. IEEE.

\bibitem[\protect\citename{Vaswani \bgroup et al.\egroup }2017]{vaswani2017attention}
Vaswani, Ashish, Noam Shazeer, Niki Parmar, Jakob Uszkoreit, Llion Jones, Aidan~N. Gomez, {\L}ukasz Kaiser, and Illia Polosukhin.
\newblock 2017.
\newblock Attention is all you need.
\newblock {\em Advances in neural information processing systems}, 30.

\bibitem[\protect\citename{Vidivelli, Ramachandran, and Dharunbalaji}2024]{vidivelli2024efficiency}
Vidivelli, S., Manikandan Ramachandran, and A.~Dharunbalaji.
\newblock 2024.
\newblock Efficiency-driven custom chatbot development: Unleashing langchain, rag, and performance-optimized llm fusion.
\newblock {\em Computers, Materials \& Continua}, 80(2).

\bibitem[\protect\citename{Wang \bgroup et al.\egroup }2024]{wang2024potential}
Wang, Calvin, Joshua Ong, Chara Wang, Hannah Ong, Rebekah Cheng, and Dennis Ong.
\newblock 2024.
\newblock Potential for gpt technology to optimize future clinical decision-making using retrieval-augmented generation.
\newblock {\em Annals of Biomedical Engineering}, 52(5):1115--1118.

\bibitem[\protect\citename{Weizenbaum}1966]{eliza}
Weizenbaum, Joseph.
\newblock 1966.
\newblock Eliza—a computer program for the study of natural language communication between man and machine.
\newblock {\em Communications of the ACM}, 9(1):36–45, January.

\bibitem[\protect\citename{Zhao \bgroup et al.\egroup }2021]{pmlr-v139-zhao21c}
Zhao, Zihao, Eric Wallace, Shi Feng, Dan Klein, and Sameer Singh.
\newblock 2021.
\newblock Calibrate before use: Improving few-shot performance of language models.
\newblock In Marina Meila and Tong Zhang, editors, {\em Proceedings of the 38th International Conference on Machine Learning}, volume 139 of {\em Proceedings of Machine Learning Research}, pages 12697--12706, Online, July. PMLR.

\end{thebibliography}

\newpage

\appendix

\appendixsection{Translated tables}
\label{appendix:a}

\begin{table}[ht!]
\caption{English version of Table \ref{prompts}.}
\begin{center}
\resizebox{\columnwidth}{!}{
\begin{tabular}{ll}
\hline
\textit{prompt type}&\textit{prompt text} \\
\hline
\textit{custom prompt} & \begin{tabular}[c]{@{}l@{}}
You are Unipa-GPT, the chatbot and virtual assistant of the University of Palermo.\\
Provide a cordially and colloquially answers to the questions provided.\\
If you receive a greeting, answer by greeting and introducing yourself.\\
If you receive a question concerning the University of Palermo,\\
answer relying on the documents given to you with the question.\\
If you do not know how to answer, apologise and suggest that you\\
consult the website, do not invent answers.\\
Question: \{question\}\\
Documents: \{context\}\end{tabular} \\
\hline
\textit{condensed prompt} & \begin{tabular}[c]{@{}l@{}}
Given the following conversation and question, rephrase the next question\\
in such a way that it is a single question.\\
Conversation: \{chat\_history\}\\
Next question: \{question\}\\
Single question:\end{tabular} \\
\hline
\end{tabular}}
\end{center}
\label{prompts-en}
\end{table}

\begin{table}[ht!]
\caption{English version of Table \ref{FT-table-contributo}.}
\begin{center}
\resizebox{\columnwidth}{!}{
\begin{tabular}{cc}
\hline
\multicolumn{1}{c}{{\textit{condensed-emb}}} & {\textit{condensed-FT-emb}} \\ \hline
\multicolumn{2}{c}{\textbf{what the all-inclusive fee is}}\\ \hline
\multicolumn{1}{l}{\begin{tabular}[c]{@{}l@{}}
The all-inclusive fee for enrolment at the University of of Palermo for the academic\\
year 2023/24 are calculated according to the student's ISEE income class and the\\
benefits provided as part of the right to study university. The payment of fees and\\
the all-inclusive fee is divided into in three instalments, the first of which must\\
be paid by 30 September 2023 for enrolment in open access study courses and for\\
enrolment to years subsequent to the first. To find out the terms and amounts of\\
payment of fees and the all-inclusive fee, we advise you to consult the Teaching\\
Calendar and the Calendar and the Student Contribution Regulations in force.\\
In case of any doubts or specific questions, contact the Student secretariat directly\\
of the University of Palermo.\\
\end{tabular}} & \begin{tabular}[c]{@{}l@{}}

The all-inclusive fee is the payment of the tuition fees that students have to pay\\
at the time of university enrolment or registration. Payment is divided into three\\
instalments: the first instalment, payable during the registration or enrolment,\\
is the 40\% of the amount due; the second instalment, to be paid by 29 December\\
2023, is the 30\% of the amount due; the third instalment, to be paid by 29 February\\
2024, is the 30\% of the amount due. Students enrolling for the first time to a\\
three-year degree or a single-cycle master's degree, who graduated with a grade\\
between 95 and 100, benefit, for the first year, from a reduction of 20\% up to a\\
maximum of €350.00. Those who graduate with a grade of 100 cum laude receive a\\
50\% reduction for the first year, up to a maximum of a maximum of 500.00 euro. $[$...$]$ \\
\end{tabular} \\ \hline
\end{tabular}}
\end{center}
\label{FT-table-contributo-en}
\end{table}

\begin{table}[ht!]
\caption{English version of Table \ref{FT-table-test}.}
\begin{center}
\resizebox{\columnwidth}{!}{
\begin{tabular}{cc}
\hline
\multicolumn{1}{c}{{ \textit{custom-full}}} & { \textit{custom-FT-clear}} \\ \hline
\multicolumn{2}{c}{\textbf{is the bachelor degree course in mechanical engineering open or restricted access?}} \\ \hline
\multicolumn{1}{l}{{ \begin{tabular}[c]{@{}l@{}}

The bachelor degree course in mechanical engineering at the University of\\
Palermo is open-access, with a number of places in local programming equal\\
to 200. This is indicated in the document for the academic year 2023/2024 for\\
the mechanics curriculum. If you need further information, I suggest you to\\
consult the university website\end{tabular}}}
& {\begin{tabular}[c]{@{}l@{}}
The bachelor degree course in mechanical engineering\\
is open-access\end{tabular}}
\\ \hline
\multicolumn{2}{c}{\textbf{do i have to take a test to enter the bachelor degree course in mechanical engineering?}} \\ \hline
\multicolumn{1}{l}{{ \begin{tabular}[c]{@{}l@{}}

Yes, to enter the bachelor degree course in mechanical engineering at the\\
University of Palermo, you must take an admission test. The University's\\
Guidance and Tutoring Center (COT) organises various initiatives to help\\
you cope with entrance tests and pass them successfully, such as the\\
‘How to cope with the entrance test’ workshops and mock entrance tests.\\
In addition, preparation courses are available for the admission tests,\\
which will start on 17 July 2023 and will take place exclusively online\\
on the Microsoft Teams platform. For more information, I advise you to\\
consult the documents provided with your application.\end{tabular}}} 
& {\begin{tabular}[c]{@{}l@{}}
Yes, there are tests for entry to the bachelor degree course\\
in Mechanical Engineering.\end{tabular}}
\\ \hline
\end{tabular}}
\end{center}
\label{FT-table-test-en}
\end{table}

\begin{table}[ht!]
\caption{English version of Table \ref{suggestions-table}}
\begin{center}
\resizebox{\columnwidth}{!}{
\begin{tabular}{cl}
\hline
\multicolumn{1}{c}{{ \textit{condensed-clear}}} & \multicolumn{1}{c}{{ \textit{custom-FT-clear}}} \\ \hline
\multicolumn{2}{c}{\textbf{which three-year course do you recommend I attend to then go on to the master's degree in aerospace engineering?}} \\ \hline
\multicolumn{1}{l}{{ \begin{tabular}[c]{@{}l@{}}
To move on to the master's degree in Aerospace Engineering at the\\
University of Palermo, the recommended three-year course is the\\
Bachelor's degree in Mechanical Engineering with Aeronautical curriculum.\\
I advise you to consult the study plan for the three-year degree course\\
and contact the Department of Engineering for further information.
\end{tabular}}} & { \begin{tabular}[c]{@{}l@{}}
I recommend that you attend the three-year degree course in Mechanical\\
Engineering and then move on to the master's degree course in Aerospace\\
Engineering. The three-year course will provide you with the necessary\\
foundations for the master's degree.
\end{tabular}} \\ \hline
\end{tabular}}
\end{center}
\label{suggestions-table-en}
\end{table}

\begin{table}[ht!]
\caption{English version of Table \ref{med-table}.}
\begin{center}
\resizebox{\columnwidth}{!}{
\begin{tabular}{cll}
\hline
\multicolumn{1}{c}{{\textit{condensed-emb}}} & \multicolumn{1}{c}{{\textit{condensed-full}}} & \multicolumn{1}{c}{\textit{pure fine-tuned model}} \\ \hline
\multicolumn{3}{c}{{\textbf{first-year medicine course exam list}}} \\ \hline
\multicolumn{1}{l}{{\begin{tabular}[c]{@{}l@{}}

The exams for the first year of the medicine\\
course at the University of Palermo for the\\
academic year 2023/24 are not specified in the\\
documents provided. I suggest you consult the\\
call for applications for more information.\\
Furthermore, as far as enrolment is concerned,\\
open access courses can be enrolled without\\
late fees from 1 August {[}...{]}\end{tabular}}} 
& \multicolumn{1}{l}{{\begin{tabular}[c]{@{}l@{}}
I am sorry, but I have not found any information\\
regarding the first-year exams of the medicine\\
course at the University of Palermo for the\\
academic year 2023/24. I advise you to consult\\
the call for applications for more information.\end{tabular}}}
& \begin{tabular}[c]{@{}l@{}}
The list of exams for the first year of the\\
degree course in Medicine at the University\\
of Palermo includes the following exams: Human\\
Anatomy with Exercises, Applied Biology, English,\\
Biological Chemistry with Elements of Molecular\\
Biology, Physics, Mathematics, Statistics and\\
Computer Science, Human Histology with\\
Elements of Cell Biology, Human Physiology with\\
Exercises, and General Pathology.\end{tabular} \\ \hline

\multicolumn{3}{c}{{\textbf{what are the subjects of the first year of medicine}}} \\ \hline
\multicolumn{1}{l}{{\begin{tabular}[c]{@{}l@{}}
The first-year subjects of the medical course\\
at the University of Palermo for the 2023/24\\
academic year are: Physics I, Mathematics I and\\
General and Organic Chemistry C.I. Instead, for\\
the degree course in Biological Sciences, the\\
first-year subjects are: Mathematics with\\
Exercises, Cytology and Histology with Exercises,\\
General Chemistry with Exercises and General\\
Botany and Histology with Exercises.
\end{tabular}}} & \multicolumn{1}{l}{{\begin{tabular}[c]{@{}l@{}}
The first-year subjects of the medical course\\
at the University of Palermo for the academic\\
year 2023/24 are: Physics I, Mathematics I,\\
General and Organic Chemistry C.I., Organic\\
Chemistry, Biology, Embryology and Histology C.I.,\\
Genetics and Microbiology C.I., Computer Science,\\
Bioinformatics and Technical Scientific English C.I.\\
I advise you to consult the documents provided\\
for more information.
\end{tabular}}} & \begin{tabular}[c]{@{}l@{}}
The first-year medical subjects are: Human\\
Anatomy, Histology, Embryology, Biochemistry,\\
Cell Biology, Genetics, Physics, Chemistry,\\
English, Methodology of Intellectual Work,\\
and Culture and Values in Healthcare.
\end{tabular}\\ \hline
\end{tabular}}
\end{center}
\label{med-table-en}
\end{table}

\begin{table}[ht!]
\caption{English version of Table \ref{prompt-sharper}.}
\begin{center}
\resizebox{\columnwidth}{!}{
\begin{tabular}{ll}
\hline
\textit{prompt type}&\textit{prompt text} \\
\hline
\textit{custom prompt} & \begin{tabular}[c]{@{}l@{}}

I am Unipa-GPT, a chatbot and virtual assistant of the University of Palermo,\\
who responds cordially and in a colloquial manner.\\
To greetings, answer by greeting and introducing yourself;\\
Answer the question with the words "Answer: "\\
Remember that the rector of the university is Professor Massimo Midiri.\\
If the question concerns the University of Palermo, answer on the basis of\\
the information and provide the links associated with it;\\
If you do not know how to answer the question, answer by saying that\\
you are an artificial intelligence that still has a lot to learn and\\
suggest that you go to https://www.unipa.it/, do not invent answers.\\
Question: \{question\}\\
Documents: \{context\}\end{tabular} \\
\hline
\textit{condensed prompt} & \begin{tabular}[c]{@{}l@{}}
Given the following conversation and question, rephrase the next question\\
in such a way that it is a single question.\\
Conversation: \{chat\_history\}\\
Next question: \{question\}\\
Single question:\end{tabular} \\
\hline
\end{tabular}}
\end{center}
\label{prompt-sharper-en}
\end{table}

\begin{table}[]
\caption{English version of Table \ref{goldenqa}.}
\begin{center}
\resizebox{\columnwidth}{!}{
\begin{tabular}{lll}
\hline
\textbf{ID} & \textbf{Questions} & \textbf{Answers} \\ \hline
\textbf{Q1}&
\begin{tabular}[c]{@{}l@{}}
\textit{Who is the Artificial}\\ 
\textit{Intelligence 1 professor}\\ 
\textit{for Computer Engineering}\\ 
\textit{Master degree course?}\\ \end{tabular} &
  \begin{tabular}[c]{@{}l@{}}
The Artificial Intelligence 1 professor for the Computer Engineering\\
Master degree course is Professor Gaglio and it will be delivered\\
during the first semester. For more information go to\\
http://www.unipa.it/struttura.html?id=721\end{tabular} \\ \hline
\textbf{Q2}&
\begin{tabular}[c]{@{}l@{}}
\textit{What are the deadlines}\\ 
\textit{for enrolling in a}\\ 
\textit{degree programme?}\end{tabular} &
  \begin{tabular}[c]{@{}l@{}}

The submission of applications for the Academic Year 2023/2024 varies\\ according to the type of course. For three-year and single-cycle\\
master's degree courses, applications can be submitted from 1 August\\
to 30 September 2023, while for master's degree courses, from 1 August\\
to 30 November 2023; in both cases, payment of the first instalment of\\
tuition fees is required. For courses with local programmed access,\\
with online booking or national programmed access, enrolment must be\\
carried out by the deadlines set out in the corresponding calls for\\
application.\end{tabular} \\ \hline
\textbf{Q3}&
\begin{tabular}[c]{@{}l@{}}
\textit{How does the master's}\\  
\textit{degree in chemistry work?}\end{tabular} &
  {\begin{tabular}[c]{@{}l@{}}

The Master's degree course in Chemistry at the University of Palermo\\
is based in Palermo and lasts 2 years. The aim of the course is to train\\
graduates with a good background, enabling them to deal competently\\
and independently with both professional activities and academic\\
and industrial research. For this reason, the course is designed\\
with the aim of completing the scientific culture in the fundamental\\ chemical disciplines and introducing more advanced knowledge.\\
Therefore, students can complete their study plan by focusing on the\\ subjects that best suit their specific interests, such as the chemistry\\
of materials, the environment and cultural heritage, supramolecular\\
chemistry and the didactics of chemistry. For more information\\
go to http://www.unipa.it/struttura.html?id=766\end{tabular}} \\ \hline
\textbf{Q4}&
\begin{tabular}[c]{@{}l@{}}
\textit{hello! I'm a guy just out of high}\\ 
\textit{school who is interested in law,}\\ 
\textit{especially corporate law.}\\ 
\textit{where should i apply?}\end{tabular} &
  \begin{tabular}[c]{@{}l@{}}
Hi! The Bachelor of Business Law Consultant programme sounds like it\\
could be just the thing for you! The degree course is based in Trapani\\
and lasts 3 years and will provide you with knowledge in the fields of\\
law and business economics. Once you have completed the course you\\
will be able to carry out interdisciplinary activities requiring legal,\\
business and organisational-managerial skills. For more information\\
go to http://www.unipa.it/struttura.html?id=1557\end{tabular} \\ \hline
\textbf{Q5}&
\begin{tabular}[c]{@{}l@{}}
\textit{how can i book an appointment}\\ 
\textit{at the secretariat?}\end{tabular} &
    \begin{tabular}[c]{@{}l@{}}
You can go to the secretariat on Mondays, Wednesdays and Fridays from\\
10 a.m. to 12 noon, Tuesdays and Thursdays from 3 p.m. to 5 p.m. .\\
You can book your appointment through the SolariQ App. For more\\
information go to https://www.unipa.it/servizi/segreterie/
    \end{tabular} \\ \hline
\textbf{Q6}&
\textit{How do I pay fees?} &
  \begin{tabular}[c]{@{}l@{}}

Fees must be paid exclusively through the PAgoPA (Public Administration\\
Payments) system, which is accessed through the university portal. After\\
completing the paperwork online, you can either pay directly online via\\
the PAgoPA system or print out the payment slip and pay at a\\
PAgoPA-enabled tax office. Further information on paying via PAgoPA\\
can be found here https://immaweb.unipa.it/immaweb/public/pagamenti.\\
seam, while the Student Contribution Regulations is available here\\
https://www.unipa.it/servizi/segreterie/.content/documents/regulations\_\\
calendars/2023/5105144-def\_regulation-student-contribution-2023-24-2.pdf \end{tabular} \\ \hline
\end{tabular}}
\end{center}
\label{goldenqa-en}
\end{table}

\end{document}